\documentclass[runningheads]{llncs}
\usepackage{graphicx}
\usepackage{bm}
\usepackage{multirow}
\usepackage{tabularx}
\usepackage{bbm}
\usepackage{here}
\usepackage[utf8]{inputenc}
\usepackage{textcomp}
\usepackage{newunicodechar}
\usepackage{graphicx}
\usepackage{booktabs}
\usepackage{cite}

\usepackage[accsupp]{axessibility}  
\newcommand{\bhline}[1]{\noalign{\hrule height #1}}

\begin{document}
\title{Attention Lattice Adapter: Visual Explanation\\Generation for Visual Foundation Models}
\titlerunning{Attention Lattice Adapter: Visual Explanation Generation}
\author{Shinnosuke Hirano \and
Yuiga Wada \and
Tsumugi Iida \and Komei Sugiura}
\authorrunning{S. Hirano et al.}
\institute{Keio University, Japan \\
\email{\{shinhirano, yuiga, tiida, komei.sugiura\}@keio.jp}\\
}
\maketitle              
\vspace{-5mm}
\begin{abstract}
In this study, we consider the problem of generating visual explanations in visual foundation models.
Numerous methods have been proposed for this purpose; however, they often cannot be applied to complex models due to their lack of adaptability.
To overcome these limitations, we propose a novel explanation generation method in visual foundation models that is aimed at both generating explanations and partially updating model parameters to enhance interpretability.
Our approach introduces two novel mechanisms: Attention Lattice Adapter (ALA) and Alternating Epoch Architect (AEA). ALA mechanism simplifies the process by eliminating the need for manual layer selection, thus enhancing the model's adaptability and interpretability. 
Moreover, the AEA mechanism, which updates ALA's parameters every other epoch, effectively addresses the common issue of overly small attention regions.
We evaluated our method on two benchmark datasets, CUB-200-2011 and ImageNet-S. 
Our results showed that our method outperformed the baseline methods in terms of mean intersection over union (IoU), insertion score, deletion score, and insertion-deletion score on both the CUB-200-2011 and ImageNet-S datasets. 
Notably, our best model achieved a 53.2-point improvement in mean IoU on the CUB-200-2011 dataset compared with the baselines.

\end{abstract}
\vspace{-8mm}
\section{Introduction}
\label{sec:intro}
\vspace{-2mm}

\begin{figure}[t]
    \centering
    \centering
    \includegraphics[height=70mm, trim=20pt 0pt 21pt 0pt, clip=true]{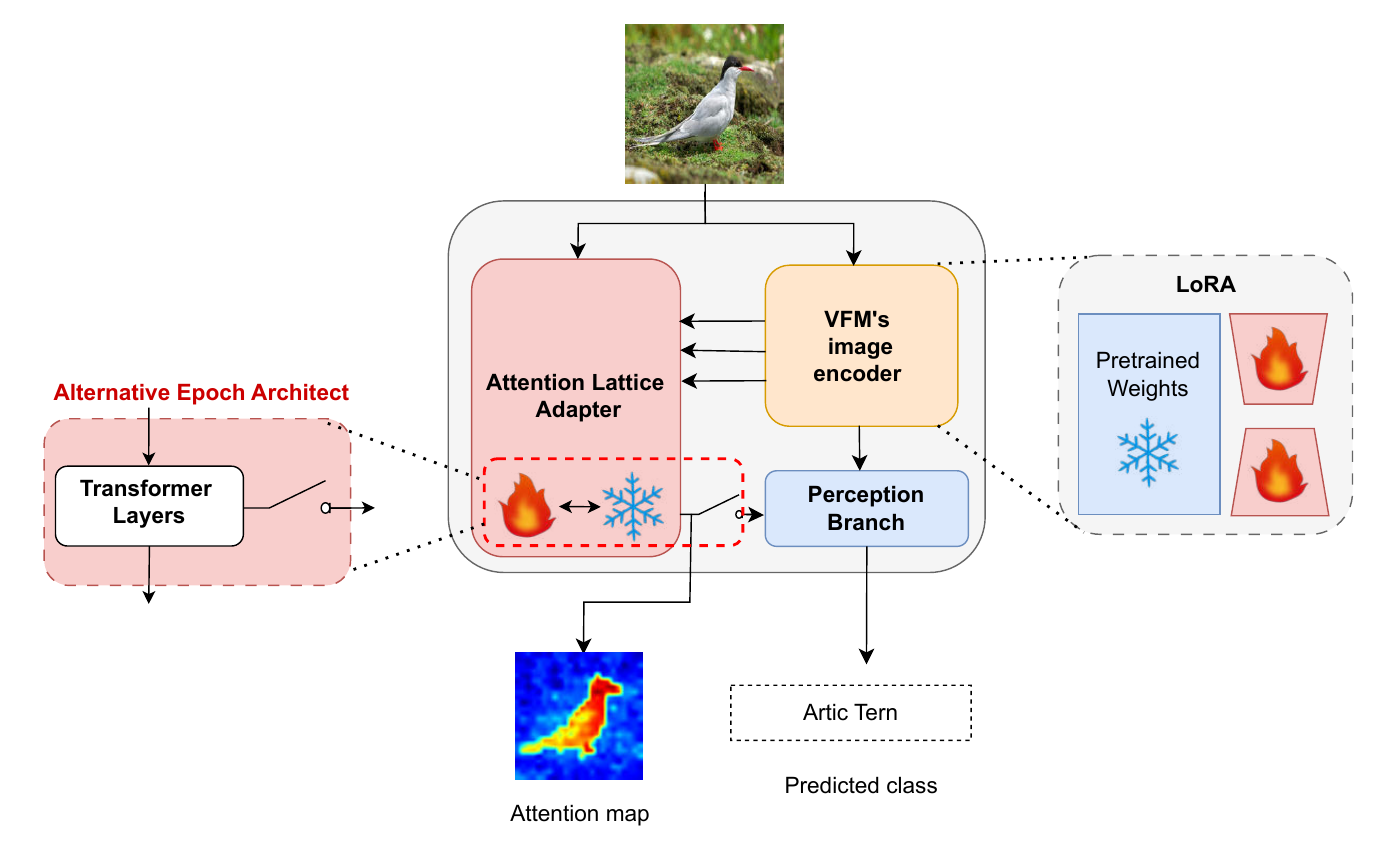}
    \vspace{-5mm}
    \caption{An overview of our method. Our method integrates an adapter into the vision foundation model (VFM) and employs Low-Rank Adaptation (LoRA) for partial training of the VFM. This approach enhances the interpretability of the generated attention maps.  Specifically, Attention Lattice Adapter (ALA) extracts multiple features from the image encoder to generate attention maps. Subsequently, the perception branch (PB) utilizes these attention maps along with the output from the image encoder to generate the final prediction probabilities.  To mitigate the common issue of excessively small attention regions, we introduce Alternating Epoch Architect (AEA), which stops updating the parameters of ALA every other epoch.
    }
    \label{eye_catch}
\end{figure}

Providing explanations for deep neural networks (DNNs) is crucial for elucidating the complex mechanisms of these models and to enhance their transparency \cite{shrikumar17alearningimportantfeatures, ribeiro2016lime}.
Although DNNs are growing in prevalence in various applications, their lack of interpretability poses significant challenges. These models, characterized by their deep and intricate architectures, often function as black boxes, with their internal prediction processes remaining largely uninterpretable. 
This lack of clarity hinders the understanding of the basis for their classifications and raises concerns about the reliability of their inferences.
Consequently, in numerous studies, researchers have proposed both model-agnostic and model-specific methods for generating reliable explanations with the aim, to enhance the reliability of these models \cite{zhou2016cvpr, Selvaraju2017gradcam ,sundararajan2017integratedgrad, srinivas2019full, bach2015layerwiserelevancepropagationlrp}.

Model-agnostic methods (e.g.\cite{Selvaraju2017gradcam}) have been successfully applied to various models; however, they sometimes struggle with complex model structures and erroneously focus on irrelevant areas.
By contrast, model-specific approaches such as Attention Branch Network (ABN) \cite{fukui2019attention} and LABN \cite{Iida2022lambdaattetnionbranchnetworkslabn} have been proposed. 
These methods incorporate specialized modules into the model's architecture to enhance interpretability through targeted explanations. 
However, these methods have a notable limitation: to manually select a single layer to extract the best features for explanation, human intervention is required.
Furthermore, as we demonstrate in Section \ref{sec:qualitative}, both model-agnostic and model-specific methods \cite{bach2015layerwiserelevancepropagationlrp, sundararajan2017axiomatic, Selvaraju2017gradcam, springenberg2015striving, wangScoreCAMScoreWeightedVisual2020} sometimes generate insufficient attention maps with overly small attention regions.
These shortcomings substantially reduce the clarity and practical utility of the generated explanations.

To alleviate these issues, in this study, we propose a method for generating explanations by introducing Attention Lattice Adapter (ALA) to a VFM.
Moreover, we introduce Alternating Epoch Architect (AEA) to effectively address the common issue of overly small attention regions.
In contrast to the ABN family \cite{fukui2019attention, Iida2022lambdaattetnionbranchnetworkslabn, magassouba2021ponnet}, our method adaptively generates an effective attention map without requiring the manual selection of a specific layer for feature extraction.
Fig. \ref{eye_catch} presents an overview of our method. 
As demonstrated in this figure, our model generates heat maps that highlight areas of interest, thereby elucidating the rationale behind the models' predictions.
Our approach, which incorporates ALA and AEA approaches, significantly enhances the interpretability of complex model predictions. This enhancement contributes to a deep understanding of the underlying mechanisms driving these predictions.

The main contributions of this study are summarized as follows:
\begin{itemize}
    \item[$\bullet$] We introduce Attention Lattice Adapter (ALA), which utilizes multiple features to enhance interpretability.
    \item[$\bullet$] We introduce Alternation Epoch Architect (AEA), which is a mechanism designed to constrain the size of attention regions by stopping updating the parameters of ALA every other epoch.
    \item[$\bullet$] Our method outperformed the baseline methods in terms of mean IoU, insertion score, deletion score, and insertion-deletion score on both the CUB-200-2011 and ImageNet-S datasets.
\end{itemize}

\vspace{-2mm}
\section{Related Work}
\vspace{-2mm}
\label{sec:related}

Numerous studies have been conducted on generating visual explanations for DNNs 
\cite{pan2021iared, wangScoreCAMScoreWeightedVisual2020, binderLayerwiseRelevancePropagation2016, cheferTransformerInterpretabilityAttention2021, Selvaraju2017gradcam, fukui2019attention}.
Several surveys offer overviews of various methods for generating explanations, along with standard evaluation metrics and datasets. These surveys also categorize and compare these approaches
\cite{Ding2022explainabilitysurvey, Joshi2021reviewonexplainabilitysurvey, zhang2021surveyonneuralnetworkinterpret}.
Methods for generating explanations in machine learning can be broadly divided into three categories: backpropagation-based methods (BP), perturbation-based methods (PER), and branch-integration-based methods (BIB). 

Because our method aims at both generating explanations and altering models to improve interpretability, it differs from existing methods that merely output explanations, including BP, PER, and others such as 
\cite{gandelsman2024interpreting, wang2023visual}.
Moreover, unlike the approaches of 
\cite{gandelsman2024interpreting, wang2023visual}, which generate explanations for the alignment between multiple modalities, our method specifically focuses on generating visual explanations for CLIP image encoders\cite{radford2021learning}.
Furthermore, our method also differs from existing BIB approaches such as ABN and LABN, which primarily extract features for generating explanations from a single layer in that we utilize a novel ALA module to extract features across multiple layers. 

\vspace{-2mm}
\subsection{Backpropagation-based (BP)}
\vspace{-2mm}

BP methods, as represented by studies such as 
\cite{zhou2016cvpr, Selvaraju2017gradcam, sundararajan2017axiomatic, smilkov2017smoothgrad, srinivas2019full, bach2015layerwiserelevancepropagationlrp, arras2017explaininglrplstm, ali2022lrptransformer}, elucidate explanations by focusing on gradients during the backpropagation process. 
Among these, in
\cite{bach2015layerwiserelevancepropagationlrp} the authors proposed Layer-wise Relevance Propagation (LRP), which generates explanations using backpropagation from the output layer. 
Additionally, the adaptation of LRP to LSTM architectures was explored in \cite{arras2017explaininglrplstm}. Similarly, in \cite{ali2022lrptransformer} the authors proposed an LRP-based method specifically tailored for Transformer architectures.
Integrated gradients (IG) \cite{sundararajan2017axiomatic} generate explanations by integrating gradients in a formulation that adheres to the principles of sensitivity and implementation invariance.
In response to the above methods, Ismail et al. \cite{ismail2021improving} indicated that most BP contain visual noise, which results in unfaithful attention maps. Therefore, Ismail et al. introduced saliency-guided training to reduce noisy gradients used in predictions while retaining the predictive performance of the model.
Although these BP are generally model-agnostic, they often focus on irrelevant areas in the application to complex models.

\vspace{-2mm}
\subsection{Perturbation-based (PER)}
\vspace{-2mm}

PER is a category of methods that generate explanations by adding perturbations to the input and observing the subsequent changes in the model's prediction. Notable examples of PER include LIME 
\cite{ribeiro2016lime}, RISE 
\cite{Petsiuk2018rise}, and Shapley sampling 
\cite{lundberg2017shap}.
For instance, RISE generates explanations by testing the model with various randomly masked versions of the input image and analyzing the resultant outputs.
However, a significant drawback of these methods is their high computational cost. This cost mainly arises from the need to repeatedly process perturbed versions of the input through the model.
Moreover, as 
Hooker et al. \cite{hooker2019roar} highlighted, there is ambiguity about whether the observed performance degradation is caused by the removal of beneficial features by the perturbations or because the perturbed samples deviate from the training data's distribution.

\vspace{-2mm}
\subsection{Branch Integration-based (BIB)}
\vspace{-2mm}

Methods categorized under BIB utilize a branch structure with a specialized module for generating explanations. Notable examples within this category include ABN
\cite{fukui2019attention}, LABN 
\cite{Iida2022lambdaattetnionbranchnetworkslabn}, Mask A3C 
\cite{itaya2021maska3cexplanationattentionbranch}, and PonNet 
\cite{magassouba2021ponnet}. 
In contrast to model-agnostic methods categorized as BP and PER, ABN was proposed to generate high-quality explanations by integrating specialized modules for explicitly making attention maps.
The ABN family approaches are principally composed of two key components: the attention branch (AB) and the perception branch (PB). 
Within the ABN framework, the AB generates an attention map by utilizing features extracted by a feature extractor.  Subsequently, the PB employs this attention map, along with the extracted features, to make predictions. 

In several experiments, researchers have demonstrated the strong performance of ABN-like models \cite{Iida2022lambdaattetnionbranchnetworkslabn, itaya2021maska3cexplanationattentionbranch, magassouba2021ponnet}; however, these methods encounter significant challenges. 
A primary challenge is the requirement for the manual selection of an appropriate layer for feature extraction, which limits their adaptability and flexibility. 
To overcome this limitation, we introduce a novel adapter mechanism, called Adaptive Layer Adapter (ALA).
This mechanism significantly enhances the model's adaptability and eliminates the necessity for manual layer selection, thereby refining the feature extraction process.

Furthermore, the ABN family encounters another significant limitation: being trained solely on labels makes the direct computation of the loss for attention regions generally infeasible. This limitation poses a challenge in effectively controlling the size of these regions, which potentially leads to overly small attention areas that focus on only a few pixels. 
We posit that this issue arises because the AB serves as a bypass, thereby leading the model to learn only a limited portion of the image encoder's feature map as an informative representation. 
Therefore, we argue for the importance of maintaining the predominance of the PB in the loss function to achieve a balance between the PB and AB. 
To balance them, we introduce AEA, which is a novel approach that halts updates to ALA’s parameters every other epoch, thereby effectively addressing the issue of overly small attention areas.

\vspace{-3mm}
\subsection{Other Methods}
\vspace{-3mm}

Gandelsman et al.\cite{gandelsman2024interpreting} proposed a novel method for generating explanations of CLIP \cite{radford2021learning} by decomposing the image representation into the sum of contributions across individual image patches, model layers, and attention heads. 
Similarly, Wang et al. \cite{wang2023visual} proposed a unique technique for explaining CLIP, using a multi-modal information bottleneck (M2IB) approach. The M2IB framework was designed to learn latent representations that effectively compress irrelevant information while retaining crucial visual and textual features, thereby enabling more focused and relevant explanations.
Unlike the methods of Gandelsman et al. \cite{gandelsman2024interpreting} and Wang et al. \cite{wang2023visual}, which focus on generating explanations for the alignment between multiple modalities, our method is specifically tailored to produce visual explanations for image encoders \cite{radford2021learning}.

\vspace{-2mm}
\subsection{Datasets}
\vspace{-2mm}

\begin{figure}[t]
    \centering
    \begin{tabular}{cccccccc}
        \begin{minipage}{0.3\hsize}
            \centering
            \includegraphics[height=30mm]{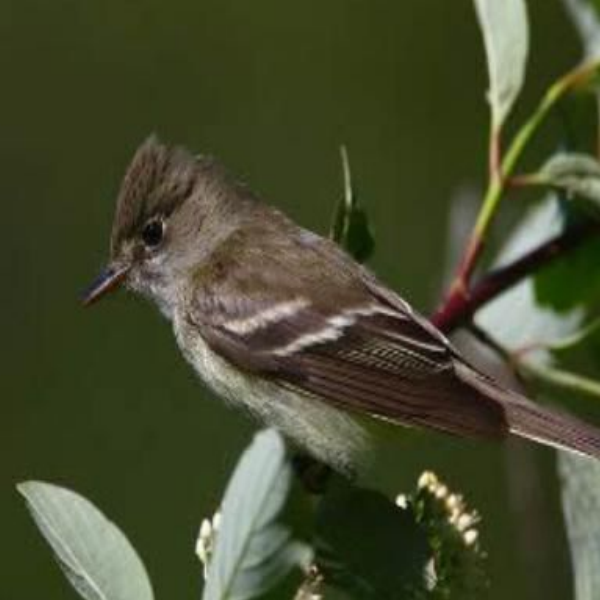}
        \end{minipage}
         &  
        \begin{minipage}{0.3\hsize}
            \centering
            \includegraphics[height=30mm]{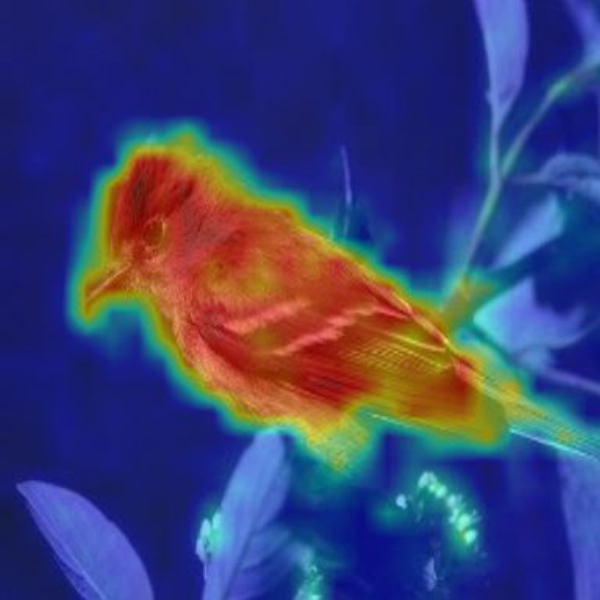}
        \end{minipage}
    \end{tabular}
\caption{Example of an input image (left) and its corresponding visual explanation (right). In this study, we focus on generating visual explanations, as illustrated in the right image, to elucidate the rationale behind the models' predictions. These visual explanations can be interpreted as masks that highlight the objects identified in the image.}
    \label{task_samples}
    \vspace{-2mm}
\end{figure}
In this field, standard datasets commonly used for generating visual explanations include ImageNet\cite{deng2009Imagenet}, CIFAR-10, and CIFAR-100. 
Additionally, domain-specific datasets such as CUB-200-2011\cite{WahCUB_200_2011} and IDRiD\cite{PORWALidrid} are often used. CUB-200-2011 consists of annotated images of birds, and comprises the class names for 200 bird species and attributes for 15 different parts. By contrast, IDRiD is a dataset designed for detecting diabetic retinopathy from retinal fundus images.

\vspace{-2mm}
\section{Problem Statement}
\label{sec:prob}
\vspace{-2mm}

We focus on the task of visualizing important regions in an image as a visual explanation of the model's predictions.
In this task, the pixels that contributed to the model's prediction should be attended.

Fig. \ref{model} shows an example of a CUB-200-2011 image. The left and right figures show the input image and visual explanation, respectively.
The input is image  $\bm{x} \in \mathbb{R} ^ {c \times h \times w}$, where $c, h$, and $w$ denote the number of channels, height and width of the input image, respectively.
The output $p(\bm{\hat{y}}) \in \mathbb{R}^C$ denotes the predicted probability for each class, where $C$ is the number of classes. 
Additionally, the importance of each pixel is obtained as a heatmap $\bm{\alpha} \in R^{h \times w}$ which is used as a visual explanation.
In this study, our specific focus is the CLIP image encoder \cite{radford2021learning}; we do not consider the CLIP text encoder.
We use mean IoU, insertion score, deletion score and insertion-deletion score as evaluation metrics for this task.

\vspace{-2mm}
\section{Proposed Method}
\label{sec:method}
\vspace{-2mm}

\begin{figure}[t]
    \centering
    \includegraphics[height=45mm]{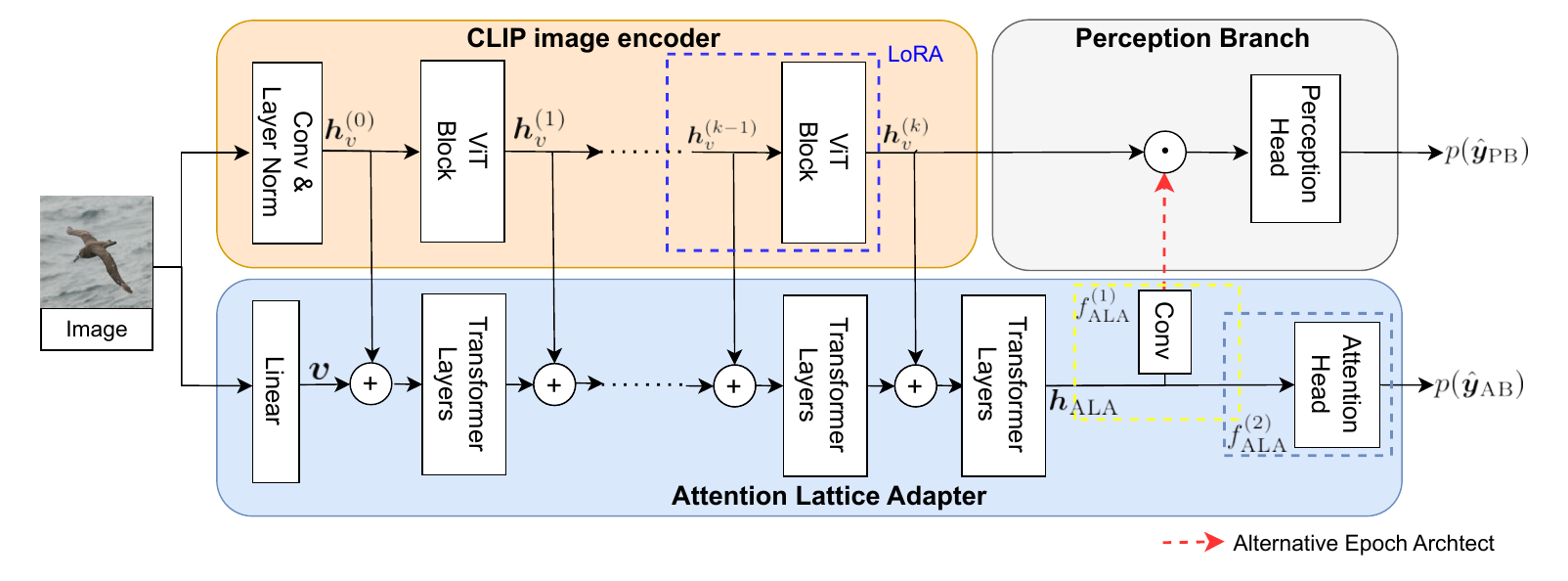} 
    \vspace{-8mm}
    \caption{Our method generates explanations by introducing Attention Lattice Adapter (ALA) to the VFM's image encoder. ALAs extract intermediate features from multiple layers of the VFM's image encoder and generate attention maps. The perception branch predicts the classification probability from the outputs of the image encoder and ALA. We apply Alternative Epoch Architect to ALA, which stops updating the parameters every other epoch to enhance interpretability.}
    \label{model}
\end{figure}

In this study, we propose a method for generating explanations by introducing Attention Lattice Adapter (ALA) to a VFM.
Inspired by \cite{fukui2019attention, xu2023side}, we use a BIB strategy in our approach.
Fig \ref{model} shows the architecture of our method.
The proposed method consists of three main modules: VFM's image encoder, ALA, and perception branch (PB). 

\vspace{-2mm}
\subsection{VFM's Image Encoder}
\vspace{-2mm}

In this study, we explore the interpretability of CLIP \cite{radford2021learning}, which is an open-vocabulary VFM pre-trained through multimodal learning. We focus on the visual component of CLIP, a detailed examination of the ViT-based CLIP image encoder, because it is expected to preserve grounded features across various resolutions.
In this module, we extract intermediate features from multiple layers of the CLIP image encoder.

Initially, we feed $\bm{x}$ into the pre-trained CLIP image encoder (ViT-B/16) composed of $k$ blocks. From this, we obtain the input to the first block $\bm{h}_{v}^{(0)}$, in addition to the intermediate features $\bm{h}_{v}^{(i)}$ at the $i$-th block.
By partially freezing the CLIP image encoder, we can improve interpretability and extract features more suitable for explanation generation while maintaining low training costs. 
Therefore, we apply LoRA \cite{hu2022lora} to the $m$-th layer of the CLIP image encoder. This approach allows us to learn only the difference in parameters through a low-rank approximation.
The parameter update formula with LoRA is expressed as follows:
\begin{equation*}
    W = W^{(0)} + BA ,
\end{equation*}
where $W$ and $W^{(0)}$ denote the updated and original pre-trained frozen weights, respectively. The matrices $A \in \mathbb{R}^{r \times d_2}$ and $B \in \mathbb{R}^{d_1 \times r}$ approximate the parameter using low-rank decomposition. In this context, $d_1$ and $d_2$ are significantly smaller than $r$, that is, $d_1 \ll r$ and $d_2 \ll r$, thereby enhancing the efficiency of the decomposition.

\vspace{-2mm}
\subsection{Attention Lattice Adapter}
\vspace{-2mm}

Generally, for the ABN family, to select an appropriate layer for feature extraction, which is critical for producing effective explanations, human intervention is required. However, this dependency considerably constrains its adaptability and flexibility. 
By contrast, our approach eliminates the need for such manual intervention by sequentially integrating multiple intermediate features within a lattice-like structure, thereby enhancing both adaptability and interpretability.

\vspace{-2mm}
\paragraph{\textbf{Former part of ALA.}} We divide $\bm{x}$ into $w_1 \times w_2$ patches. We then project these patches into visual features, $\bm{v}$, using a linear projection. 
Next, we feed $\bm{v}$ into the $l$-th transformer layer\cite{vaswani2017attention}.
At each layer $i \;(i=1,2,3,4)$, we added $\bm{h}_v^{i-1}$ to the input before the $i$-th layer. 
The output from the $l$-th transformer layer is denoted by $\bm{h}_{l}$.

\vspace{-2mm}
\paragraph{\textbf{Latter part of ALA.}} The latter part of ALA, denoted by $f_{\mathrm{ALA}}$, computes both the attention map $\bm{\alpha}$, which focuses on relevant regions, and the predicted probabilities $p(\hat{\bm{y}}_{\mathrm{ALA}}) \in \mathbb{R}^C$ for attention loss calculation. 
This approach enables the generation of attention maps closely linked to the prediction result. 
The process $f_{\mathrm{ALA}}$ is split into two subparts: $f^{(1)}_{\mathrm{ALA}}$ and $f^{(2)}_{\mathrm{ALA}}$. 

First, $f^{(1)}_{\mathrm{ALA}}$ creates an attention map using a convolution layer with $\bm{h}_{l}$ as its input. 
We could directly use $\bm{\alpha}$, that is, the output of $f^{(1)}_{\mathrm{ALA}}$; however, this approach can lead to focusing on overly small regions. 
This issue arises from the necessity to generate attention regions solely based on labels, without ground truth masks.
Consequently, computing the loss directly for these attention regions becomes a significant challenge. 
The tendency to focus on overly small regions may stem from the presence of ALA as a bypass, which leads to the learning of only a portion of the feature map as an informative representation. 
Therefore, we posit that it is crucial to maintain the predominance of PB in the loss function and to effectively balance ALA and PB.
To alleviate this issue, we introduce Alternative Epoch Architecture (AEA), which balances ALA and PB utilizing the following $\bm{h}_{\mathrm{ALA}}$ as the output of $f^{(1)}_{\mathrm{ALA}}$ in every other epoch:
\begin{equation*}
    \bm{h}_{\mathrm{ALA}} = 
    \begin{cases}
    \bm{\alpha} & (e=2n-1) \;, \\
    \mathbbm{1} & (e=2n) \; ,
    \end{cases} 
\end{equation*}
where $\mathbbm{1} \in \mathbb{R}^{h \times w}$ represents a matrix with all elements equal to 1 and $e, n$ represents the epoch number and any natural number, respectively. 
Additionally, the parameters of ALA are frozen every other epoch, thereby mitigating biased learning toward ALA and resolving the issue of overly small attention regions.

Second, the function $f^{(2)}_{\mathrm{ALA}}$ computes the predicted probabilities $p(\hat{\bm{y}}_{\mathrm{ALA}})$.
This function consists of convolutional layers, pooling layers, global average pooling, and fully connected layers. 
Specifically, $f^{(2)}_{\mathrm{ALA}}$ takes $\bm{h}_{\mathrm{ALA}}$ as input and outputs the predicted probabilities $p(\hat{\bm{y}}_{\mathrm{ALA}})$, which are used for calculating the attention loss.

\vspace{-2mm}
\subsection{Perception Branch}
\vspace{-2mm}

PB is a crucial component of our model that is responsible for generating final class predictions.
This module comprises convolutional layers, pooling layers, and fully connected layers. 
It takes $\bm{h}_{v}^{(k)}$ and $\bm{h}_{\mathrm{ALA}}$ as inputs and outputs the predicted classification probability $p(\hat{\bm{y}}_{\mathrm{PB}})$.
The operation of this module is given by the following equations:
\begin{equation*}
\begin{split}
    &p(\hat{\bm{y}}_{\mathrm{PB}}) = f_{\mathrm{PB}} (\bm{\alpha} \odot \bm{h}_{v}^{(k)}), \\
    &p(\hat{\bm{y}}_{\mathrm{ALA}}) = f_{\mathrm{ALA}}^{(1)} (\bm{h}_\mathrm{ALA}).
\end{split}
\end{equation*}
$p(\hat{\bm{y}}_{\mathrm{PB}})$ is utilized to output the prediction results. 
Given that $\bm{h}_{\mathrm{ALA}}$ represents the importance of each pixel, the Hadamard product of $\bm{h}_{v}^{(k)} \odot \bm{h}_{\mathrm{ALA}}$ effectively highlights regions crucial for prediction.
It is important to note that although $p(\hat{\bm{y}}_{\mathrm{ALA}})$ is not directly used for classification, its incorporation into the loss function can enhance the explanation quality.

Following other ABN family models \cite{fukui2019attention,Iida2022lambdaattetnionbranchnetworkslabn}, we incorporate $p(\hat{\bm{y}}_\mathrm{PB})$ and $p(\hat{\bm{y}}_\mathrm{ALA}$ into our loss function formulation. The loss function is defined as follows:
\begin{equation*}
    \mathcal{L} = \mathrm{CE}(p(\hat{\bm{y}}_{\mathrm{PB}}), \bm{y}) + \lambda \mathrm{CE}(p(\hat{\bm{y}}_{\mathrm{ALA}}), \bm{y}),
\end{equation*}
where $\bm{y}$, $\lambda$ and CE denote ground truth label, weight and cross-entropy loss function, respectively.
\section{Experimental Evaluation}
\label{sec:setup}

\subsection{Experimental Setting}

\paragraph{\textbf{Datasets.}}
We employed two distinct datasets to evaluate our model's performance in the visual explanation task: CUB-200-2011 \cite{WahCUB_200_2011} and ImageNet-S \cite{gao2022luss} datasets. The CUB-200-2011 dataset \cite{WahCUB_200_2011} is a standard dataset for our target task that comprises 11,788 images across 200 bird species classes. Each image in this dataset is accompanied by a pixel-level mask. This dataset was compiled from a list of 200 bird species sourced from an online field guide.

The ImageNet-S dataset is a modified version of the ImageNet dataset \cite{deng2009Imagenet} that is distinguished by its lack of segmentation masks. This dataset includes 1,195,741 images spanning 919 classes with pixel-level masks provided for 21,609 of these images. To construct the ImageNet-S dataset, annotators reclassified the samples from the original ImageNet dataset into their appropriate categories and annotated the corresponding masks.

For the CUB-200-2011 dataset, we used 5,000 samples for the training set, 994 for the validation set, and 5,794 for the test set. It is important to note that the CUB-200-2011 dataset’s standard partitioning comprises only training and test sets. Consequently, we kept the official test set intact and split the training set into customized training and validation sets.
Regarding the ImageNet-S dataset, we divided it into training, validation, and test sets containing 1,174,132, 9,190, and 12,419 samples, respectively. Because the official test masks were not provided, we selected 9,190 images with masks from the training set to form our validation set and used the official validation sets as our test set.
We used the training set to train our model, the validation set to tune the hyperparameters and test set to evaluate our model.

\vspace{-2mm}
\paragraph{\textbf{Baseline methods.}} 
We selected the following methods as the baselines: Layer-wise Relevance Propagation (LRP)  \cite{bach2015layerwiserelevancepropagationlrp}, integrated gradients (IG) \cite{sundararajan2017axiomatic}, guided backpropagation (Guided BP) \cite{springenberg2015striving}, Score-CAM \cite{wangScoreCAMScoreWeightedVisual2020} and Grad-CAM  \cite{Selvaraju2017gradcam}.
We selected these because they are standard methods for generating explanations. 
\vspace{-2mm}
\paragraph{\textbf{Implementation Details.}}
Table \ref{tab: parameter} summarizes our experimental setups. Note that $\#A$, $\#L$, $\#P$, and $\#d$ denote the number of attention heads, number of transformer layers in Attention Lattice Adapter, patch size and the dimensionality of the transformer, respectively.
As part of the preprocessing steps, we resized the input images to a uniform dimension of 384 $\times$ 384 pixels. During the training process, we applied flipping and cropping to the input images as data augmentation.
Furthermore, we used early stopping in our model. If no improvement was observed over five consecutive epochs on the validation set, we stopped training and then evaluated the model's performance using the test set.

\begin{table}[t]
\centering
\caption{Experimental settings for the proposed method.}
\vspace{-2mm}
\begin{tabular}{lll}
\bhline{1.25pt} 
Dataset & CUB-200-2011\cite{WahCUB_200_2011}  & ImageNet-S\cite{gao2022luss}  \\ \hline
Epoch  & 120 & 20    \\
Batch size  & 8  & 64 \\
Learning rate & $5.0 \times 10^{-5}$ & 5.0 $\times 10^{-5}$ \\
Optimizer   & \begin{tabular}{l} AdamW \\ ($\beta_1=0.9, \beta_2=0.999$) \end{tabular} & \begin{tabular}{l} AdamW \\ ($\beta_1=0.9, \beta_2=0.999$) \end{tabular}  \\ 
Weight decay  & $1.0 \times 10^{-4}$ & $1.0 \times 10^{-4}$ \\ 
Attention Lattice Adapter  & \begin{tabular}{l} $\#A$: 4 $\#L$: 8 \\ $\#P$: 16 $\#d$: 64 \end{tabular} & \begin{tabular}{l} $\#A$: 6 $\#L$: 8 \\ $\#P$:16 $\#d$: 240 \end{tabular} \\ \bhline{1.25pt}
\end{tabular}
\label{tab: parameter}
\vspace{-4mm}
\end{table}

\vspace{-2mm}
\subsection{Qualitative Results}
\vspace{-2mm}

\begin{figure}[t]
    \centering
    \begin{tabular}{cccccccc}
        \begin{minipage}{0.125\hsize}
            \centering
            \includegraphics[height=16mm, width=16mm, keepaspectratio=false]{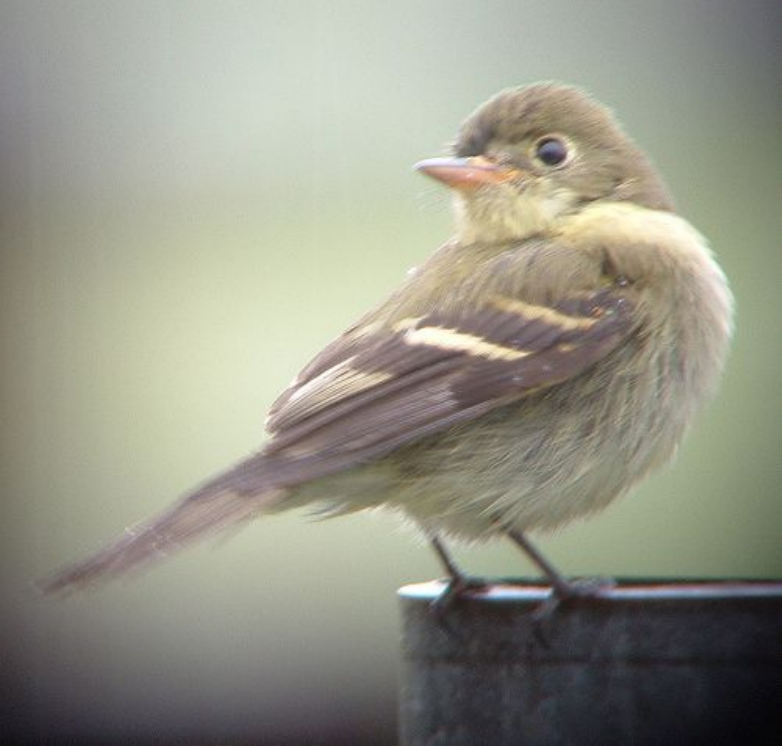}
        \end{minipage}
         &  
        \begin{minipage}{0.125\hsize}
            \centering
            \includegraphics[height=16mm, width=16mm, keepaspectratio=false]{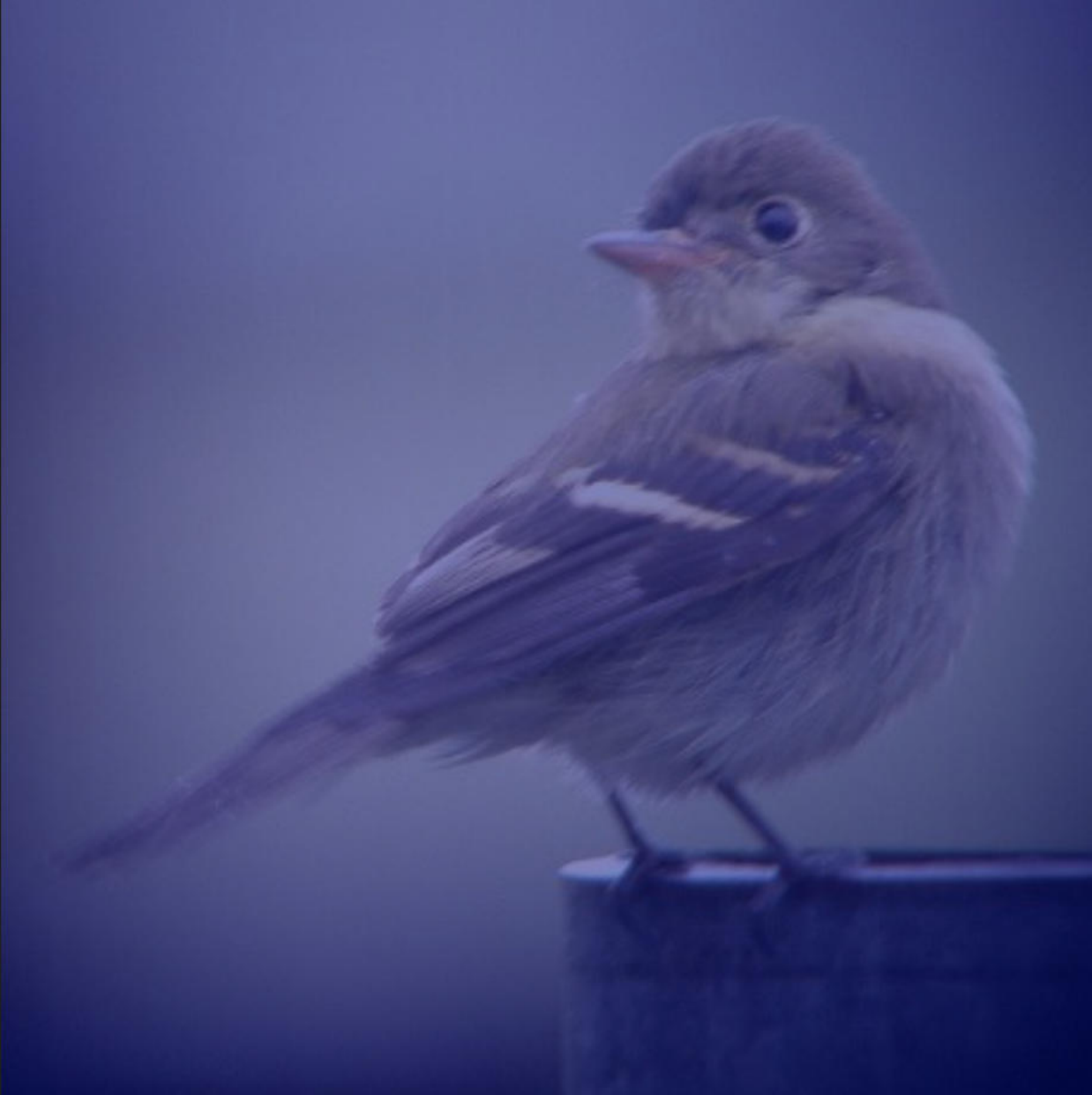}
        \end{minipage}
         & 
         \begin{minipage}{0.125\hsize}
            \centering
            \includegraphics[height=16mm, width=16mm, keepaspectratio=false]{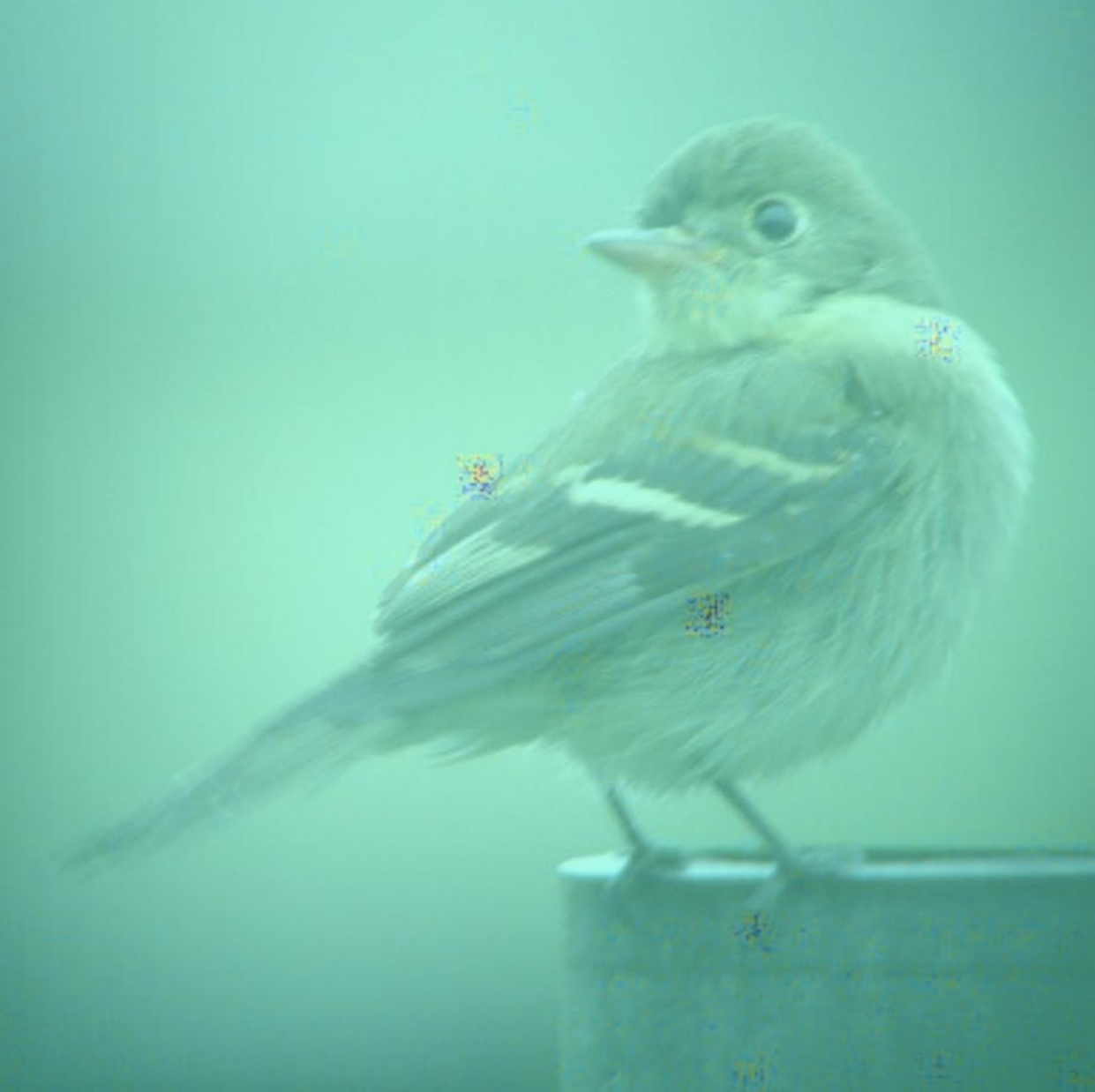}
        \end{minipage}
        &
         \begin{minipage}{0.125\hsize}
            \centering
            \includegraphics[height=16mm, width=16mm, keepaspectratio=false]{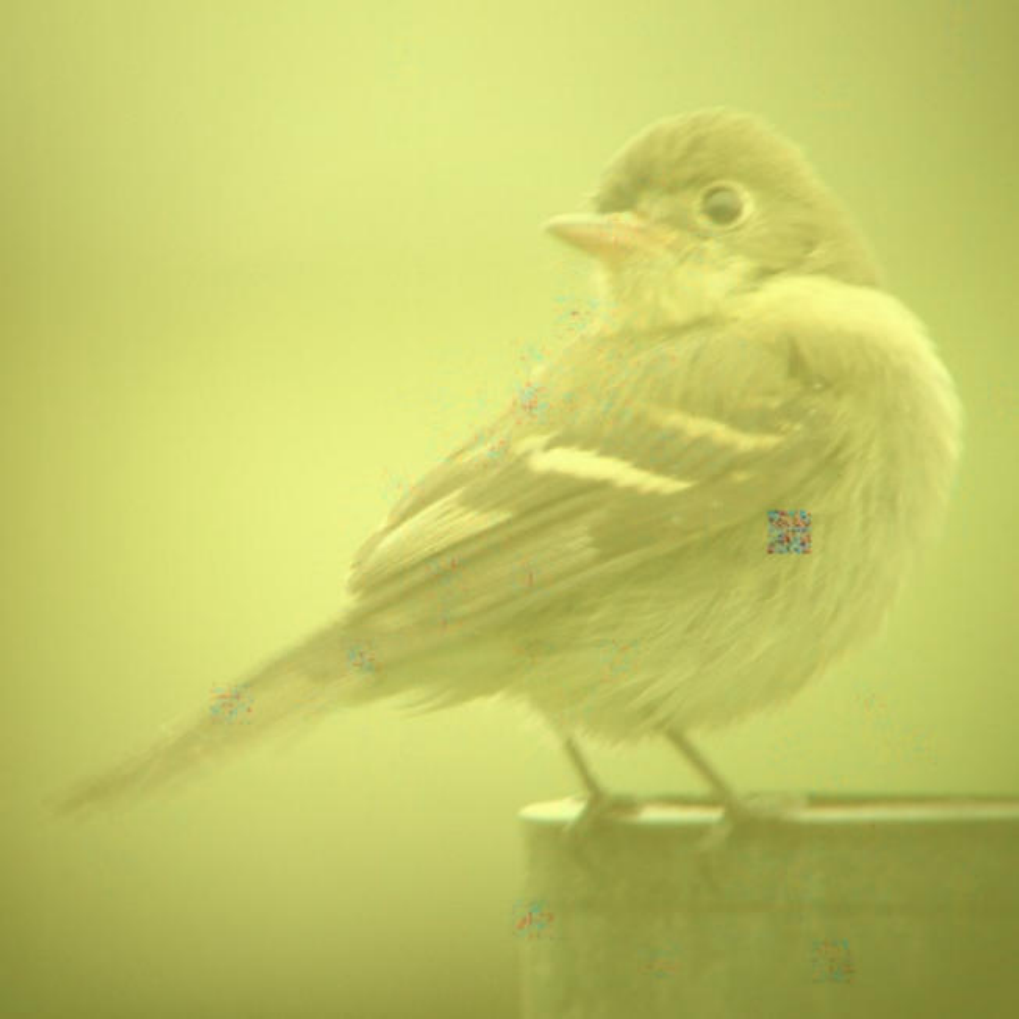}
        \end{minipage}
        &
         \begin{minipage}{0.125\hsize}
            \centering
            \includegraphics[height=16mm, width=16mm, keepaspectratio=false]{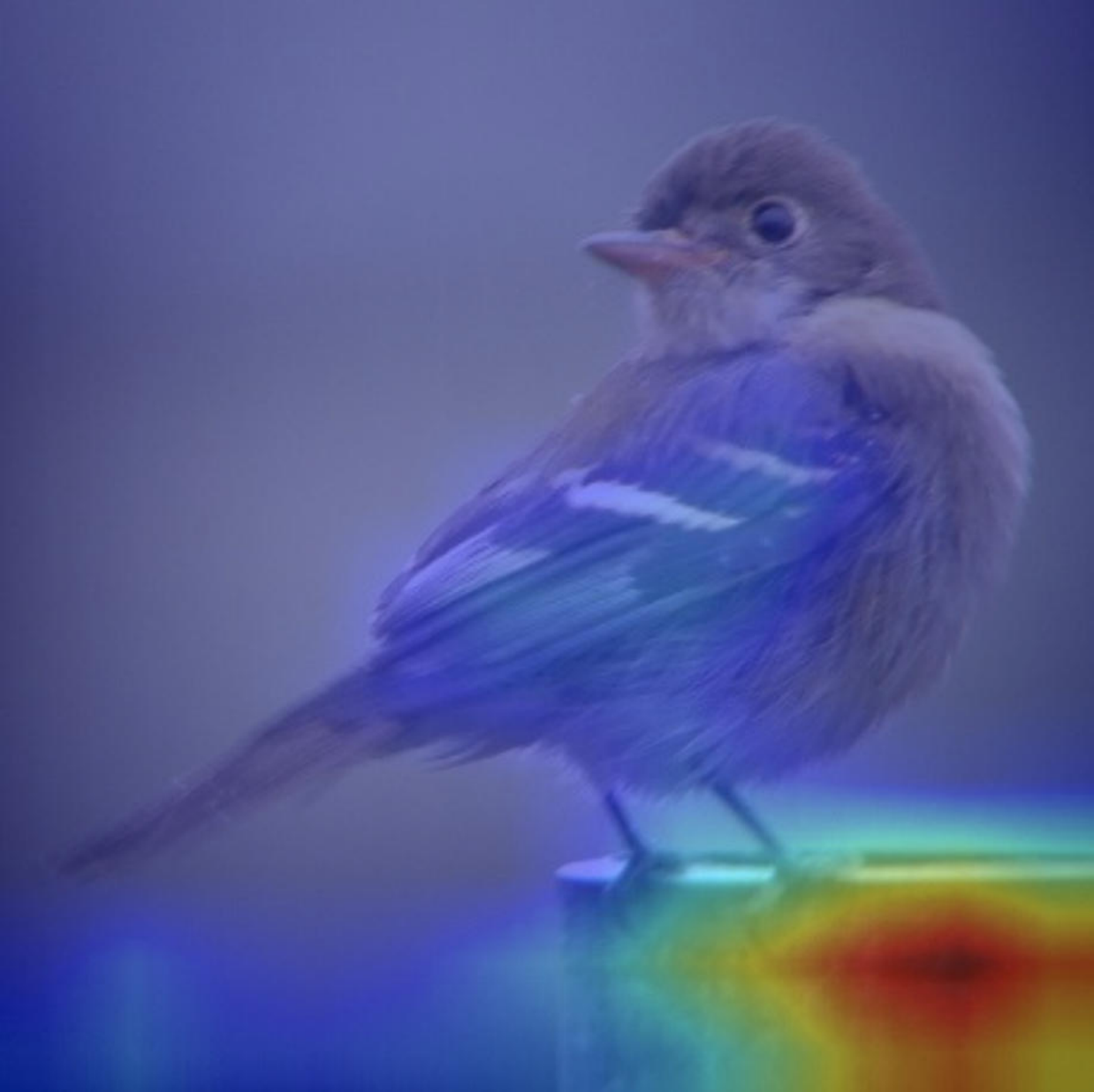}
        \end{minipage}
        &
        \begin{minipage}{0.125\hsize}
            \centering
            \includegraphics[height=16mm, width=16mm, keepaspectratio=false]{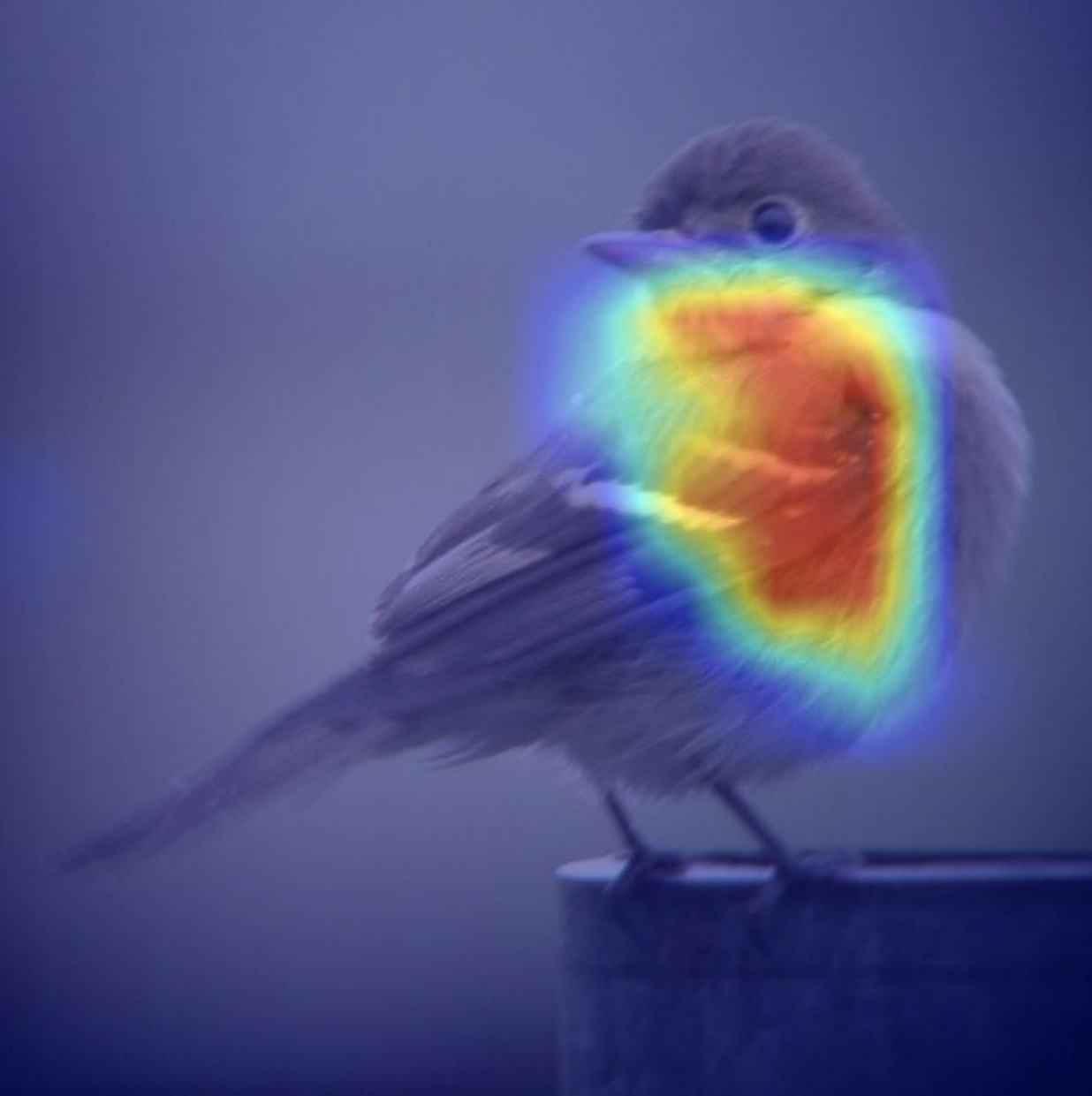}
        \end{minipage}
         \begin{minipage}{0.125\hsize}
            \centering
            \includegraphics[height=16mm, width=16mm, keepaspectratio=false]{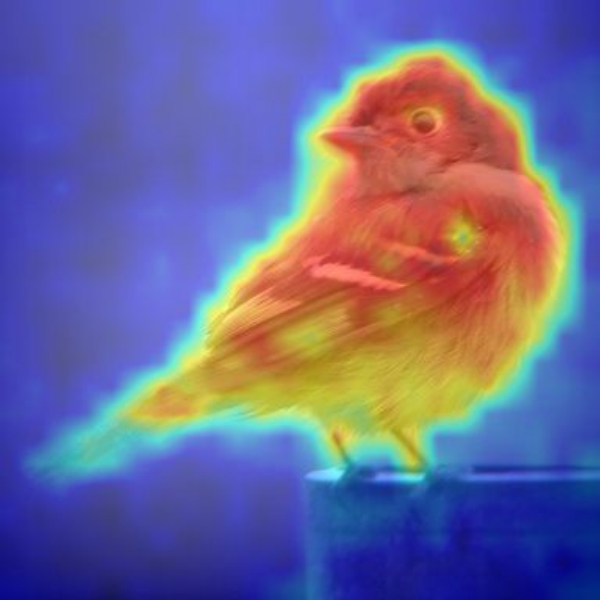}
        \end{minipage}
    \end{tabular} \\
    \begin{tabular}{cccccccc}
        \begin{minipage}{0.125\hsize}
            \centering
            \includegraphics[height=16mm, width=16mm, keepaspectratio=false]{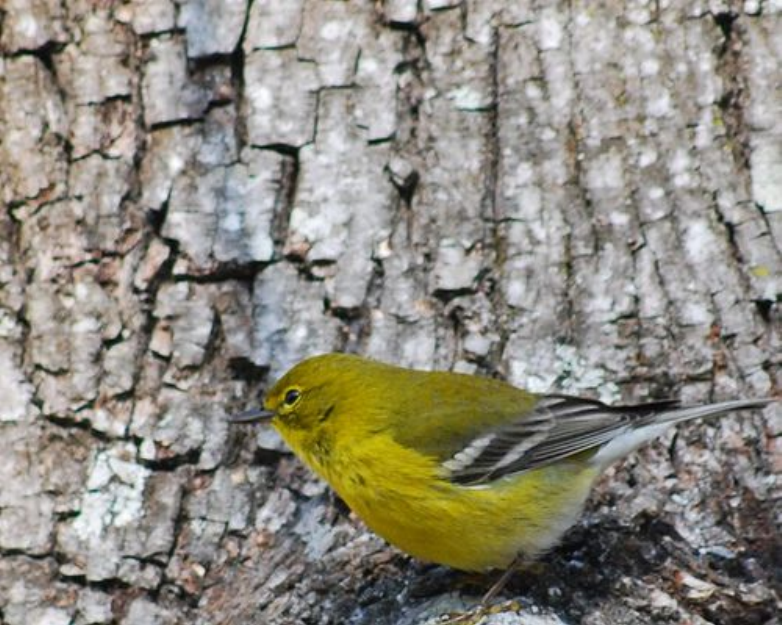}
        \end{minipage}
         &  
        \begin{minipage}{0.125\hsize}
            \centering
            \includegraphics[height=16mm, width=16mm, keepaspectratio=false]{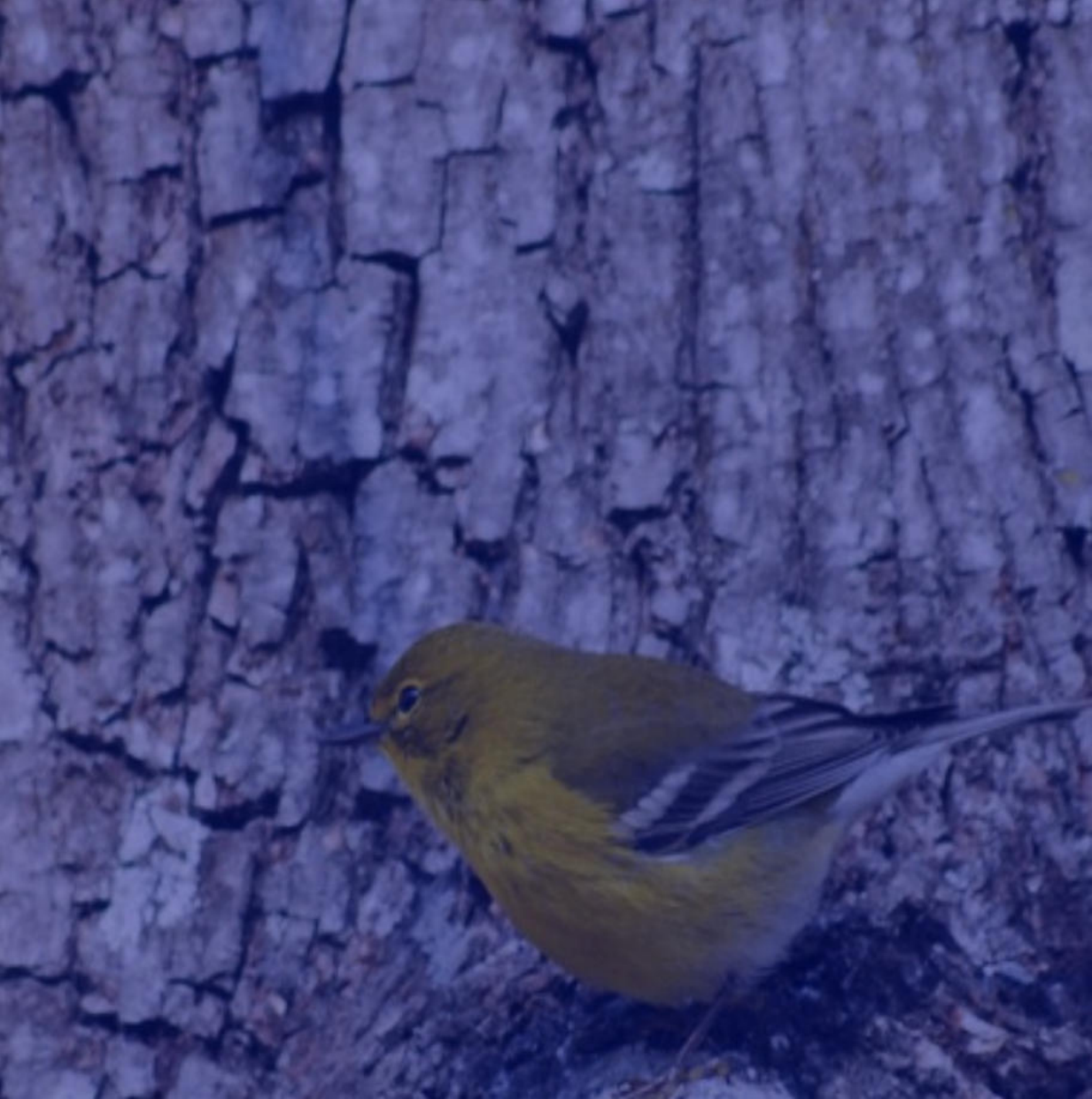}
        \end{minipage}
         & 
         \begin{minipage}{0.125\hsize}
            \centering
            \includegraphics[height=16mm, width=16mm, keepaspectratio=false]{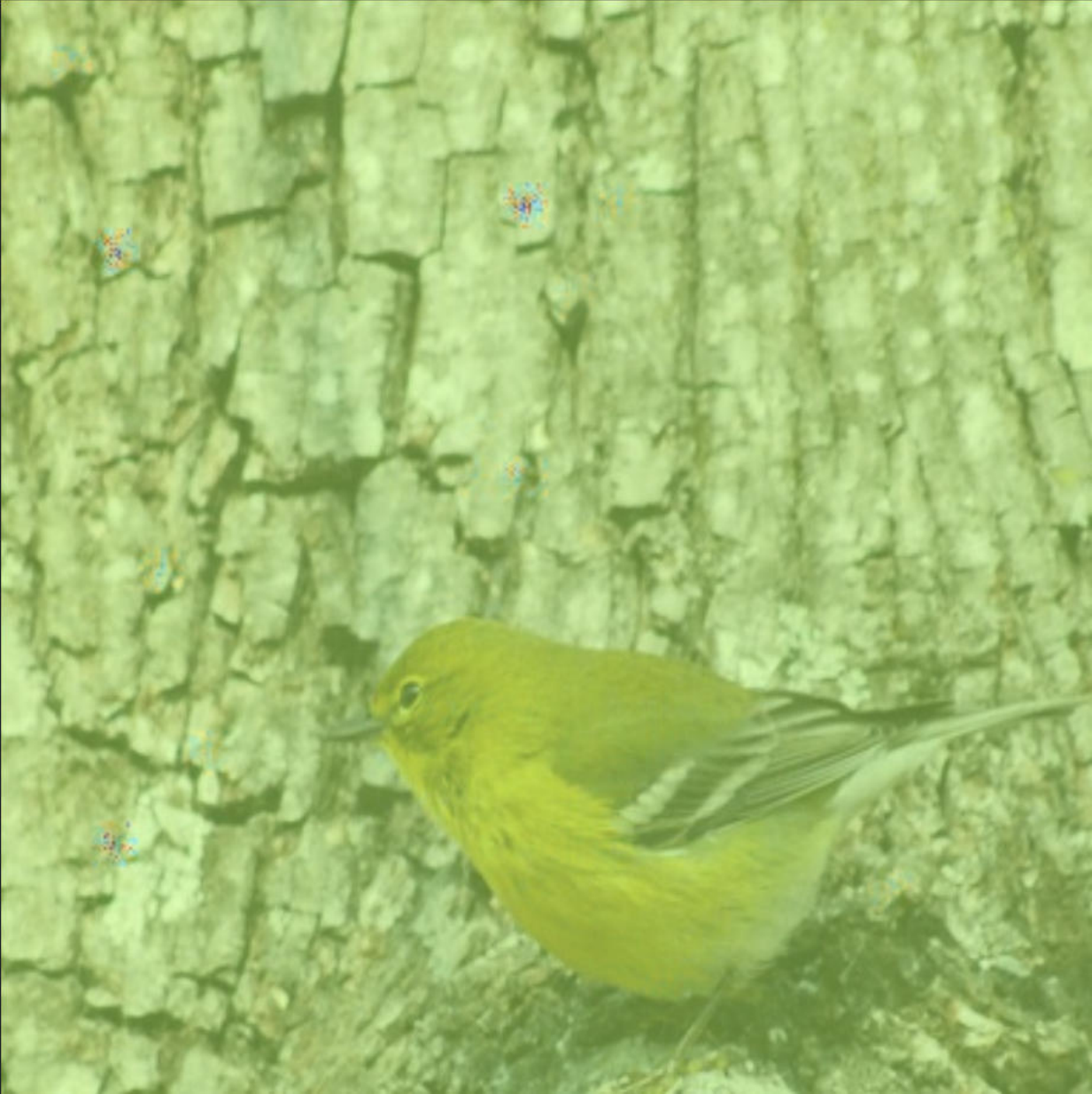}
        \end{minipage}
        &
         \begin{minipage}{0.125\hsize}
            \centering
            \includegraphics[height=16mm, width=16mm, keepaspectratio=false]{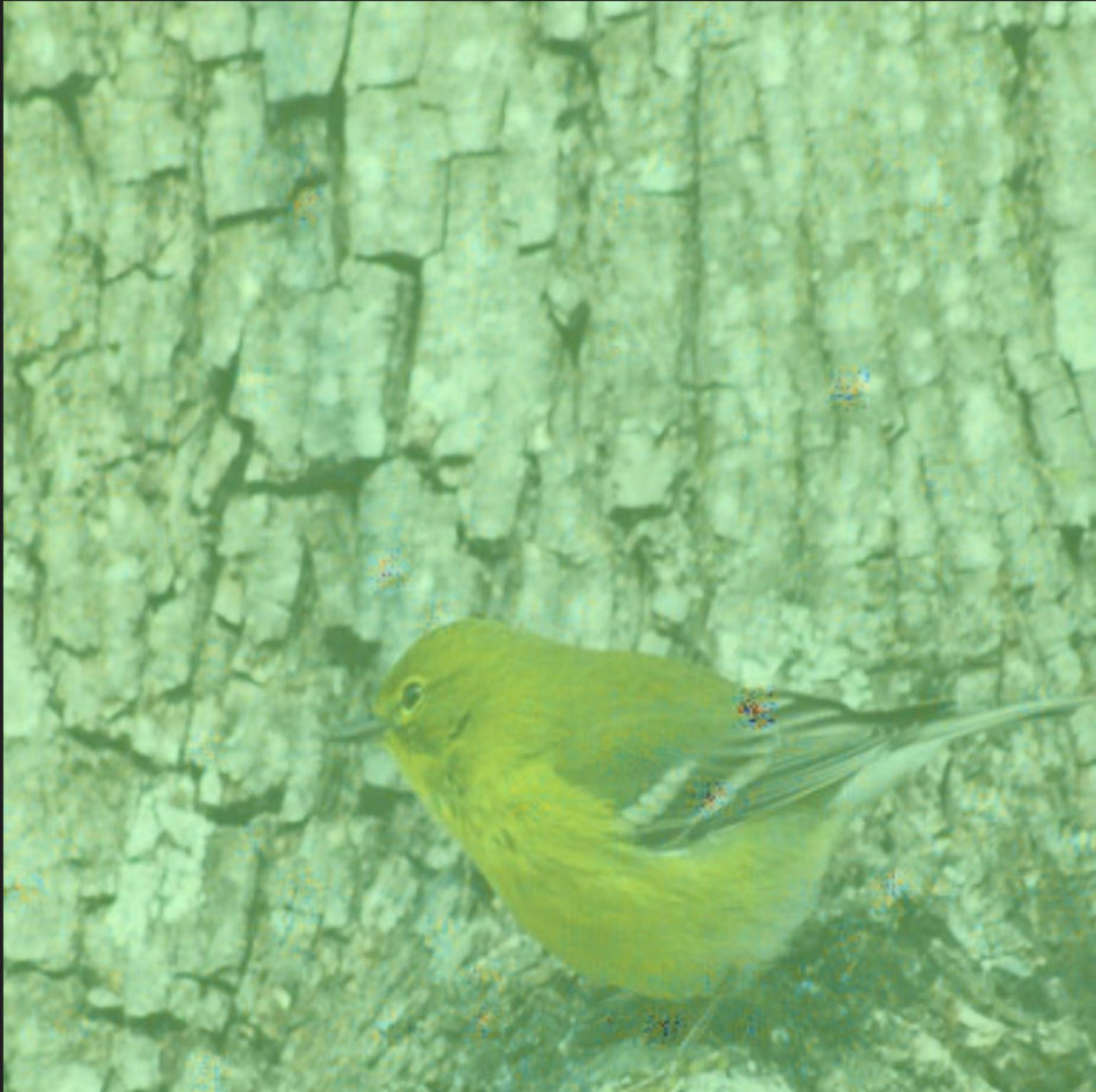}
        \end{minipage}
        &
         \begin{minipage}{0.125\hsize}
            \centering
            \includegraphics[height=16mm, width=16mm, keepaspectratio=false]{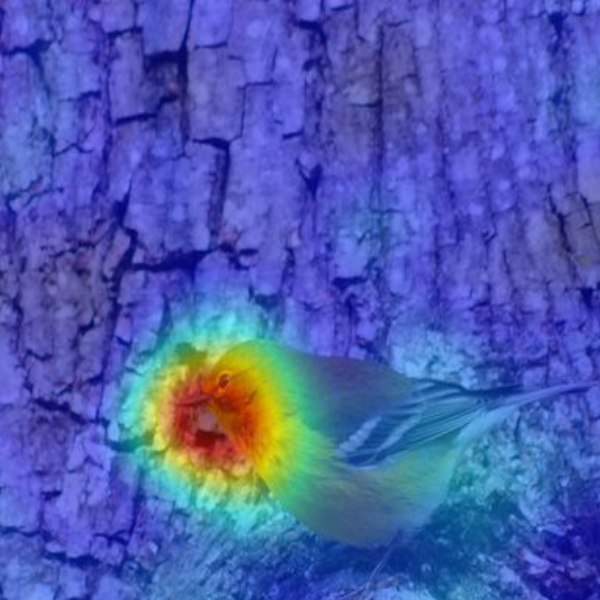}
        \end{minipage}
        &
        \begin{minipage}{0.125\hsize}
            \centering
            \includegraphics[height=16mm, width=16mm, keepaspectratio=false]{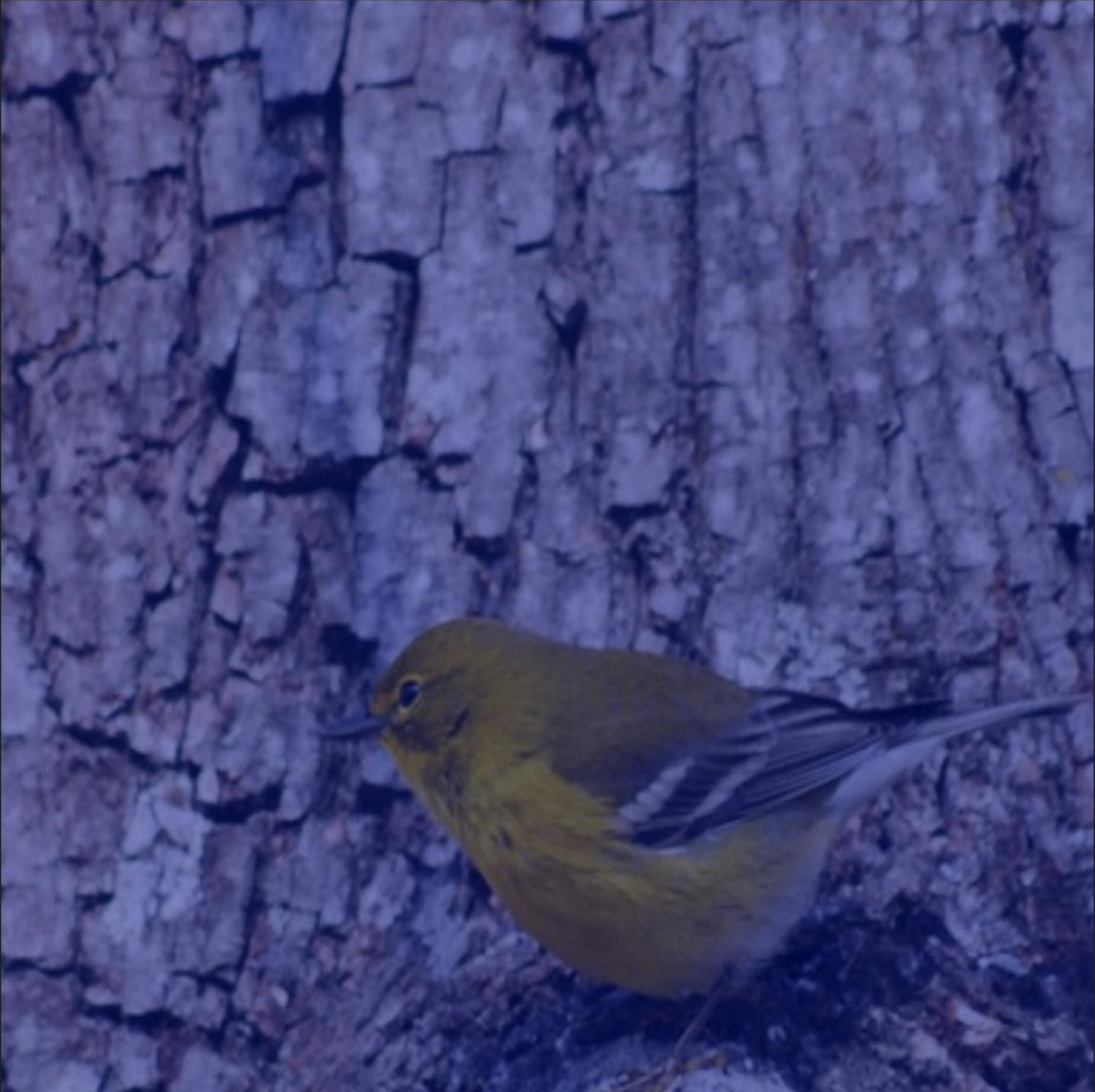}
        \end{minipage}
         \begin{minipage}{0.125\hsize}
            \centering
            \includegraphics[height=16mm, width=16mm, keepaspectratio=false]{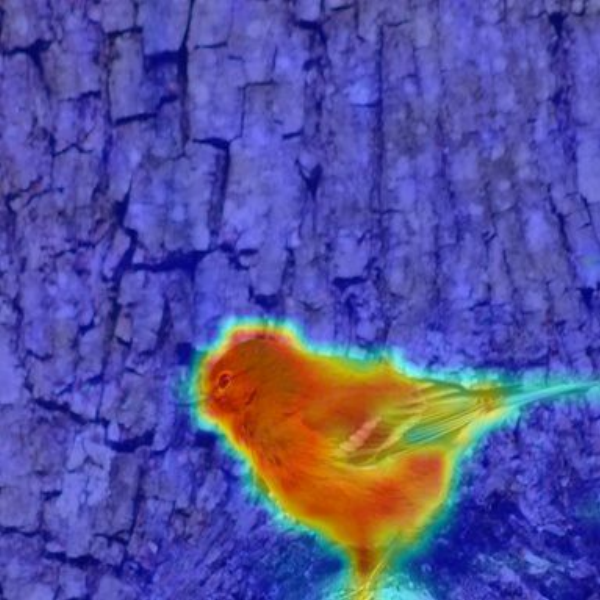}
        \end{minipage}
    \end{tabular} \
    \begin{tabular}{cccccccc}
        \begin{minipage}{0.125\hsize}
            \centering
            \includegraphics[height=16mm, width=16mm, keepaspectratio=false]{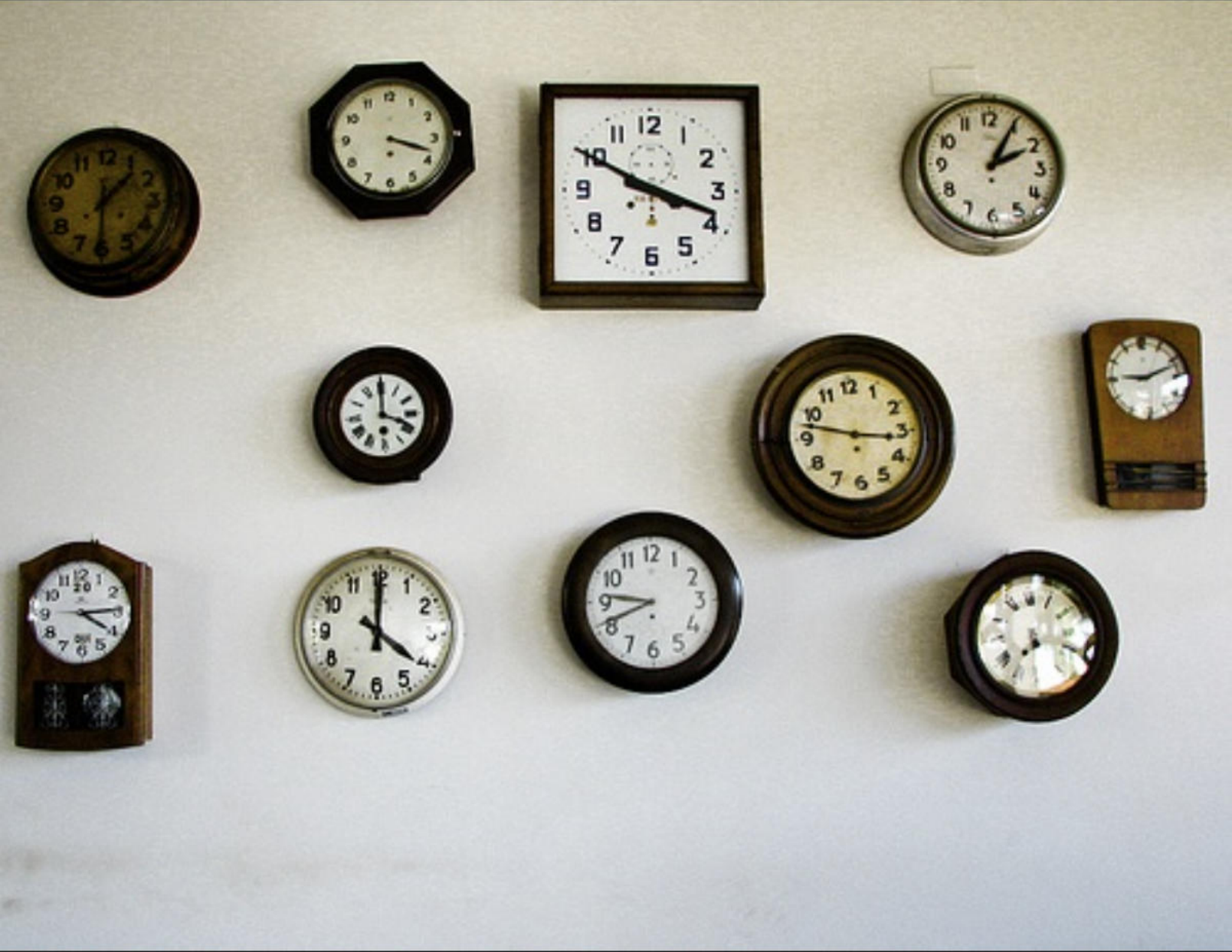}
        \end{minipage}
         &  
        \begin{minipage}{0.125\hsize}
            \centering
            \includegraphics[height=16mm, width=16mm, keepaspectratio=false]{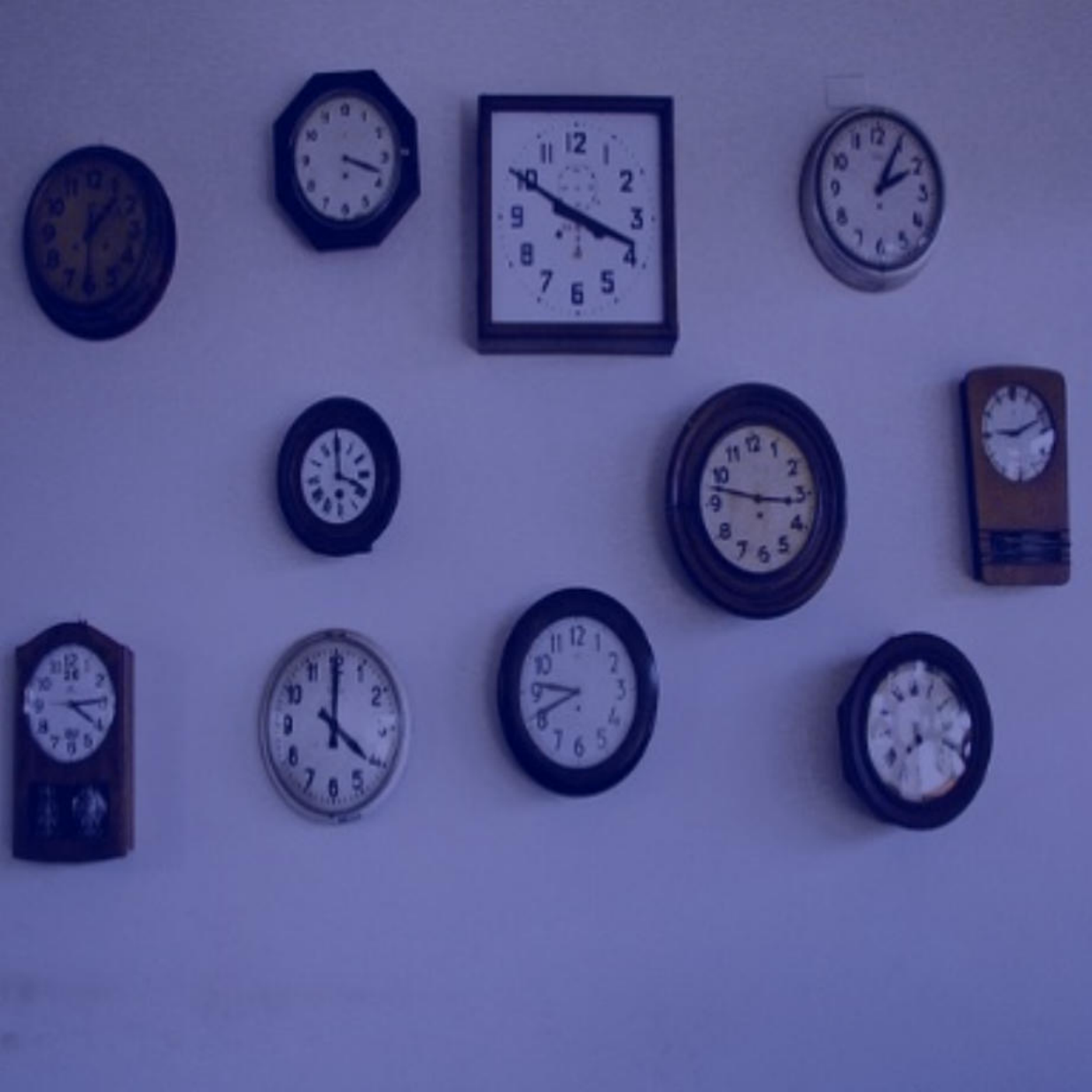}
        \end{minipage}
         & 
         \begin{minipage}{0.125\hsize}
            \centering
            \includegraphics[height=16mm, width=16mm, keepaspectratio=false]{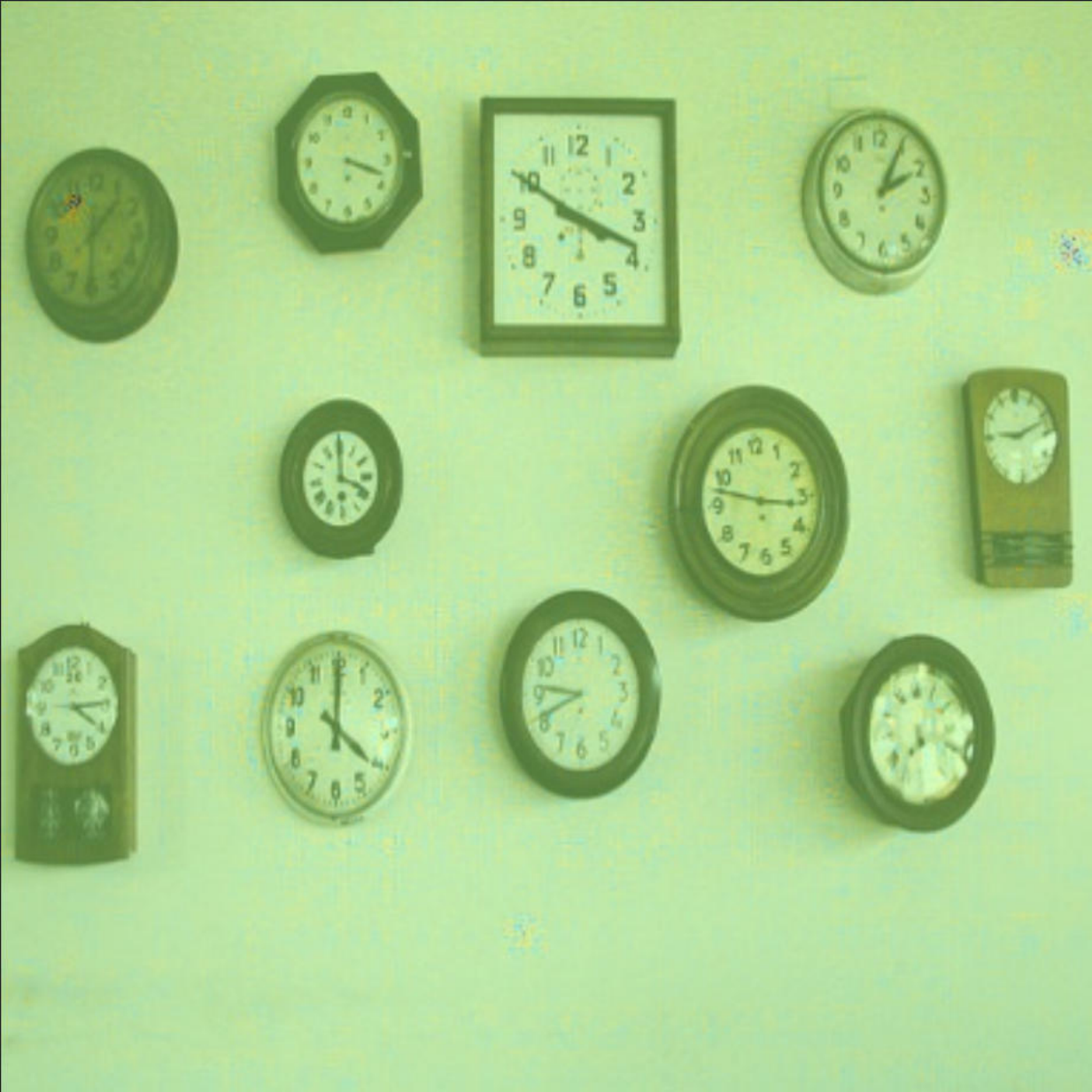}
        \end{minipage}
        &
         \begin{minipage}{0.125\hsize}
            \centering
            \includegraphics[height=16mm, width=16mm, keepaspectratio=false]{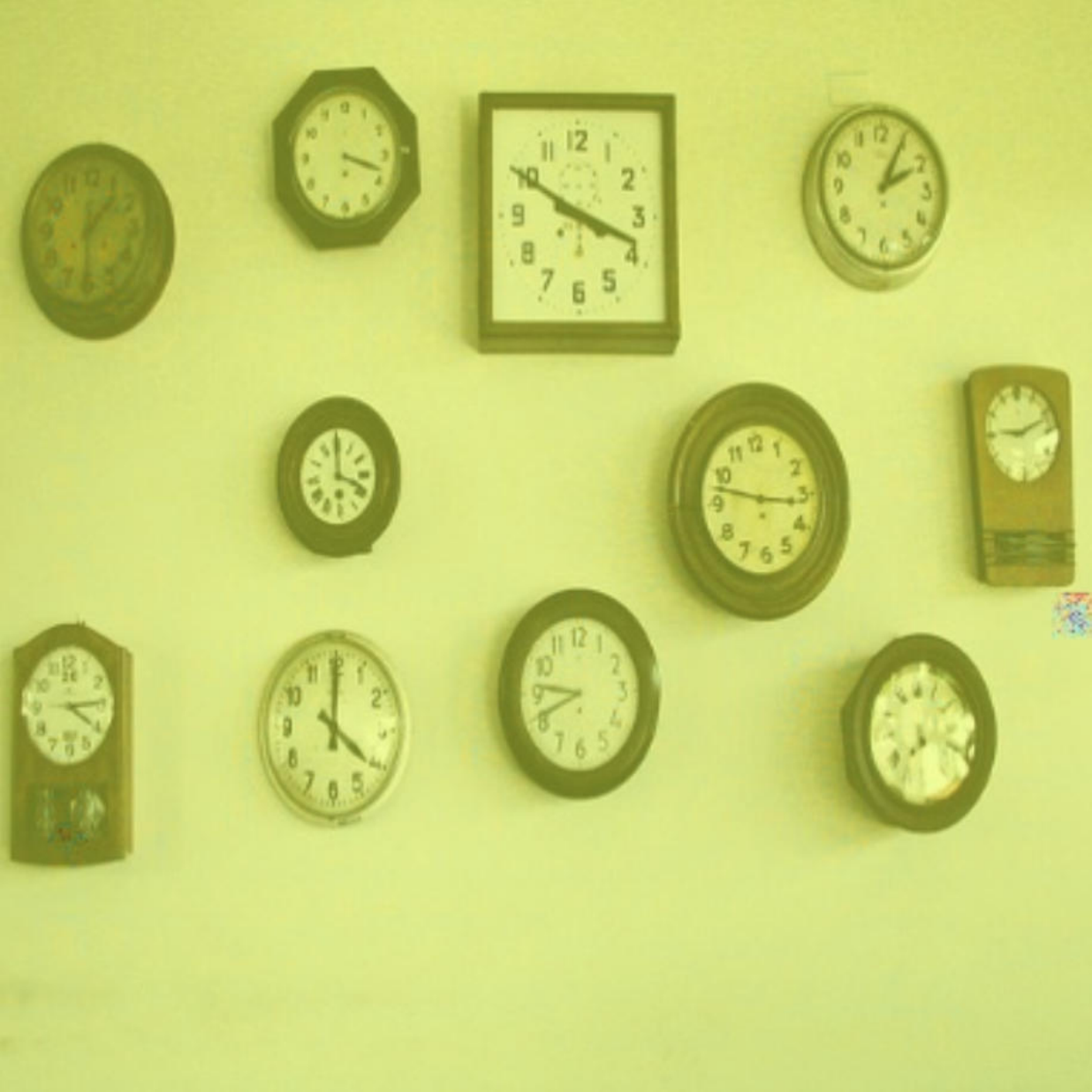}
        \end{minipage}
        &
         \begin{minipage}{0.125\hsize}
            \centering
            \includegraphics[height=16mm, width=16mm, keepaspectratio=false]{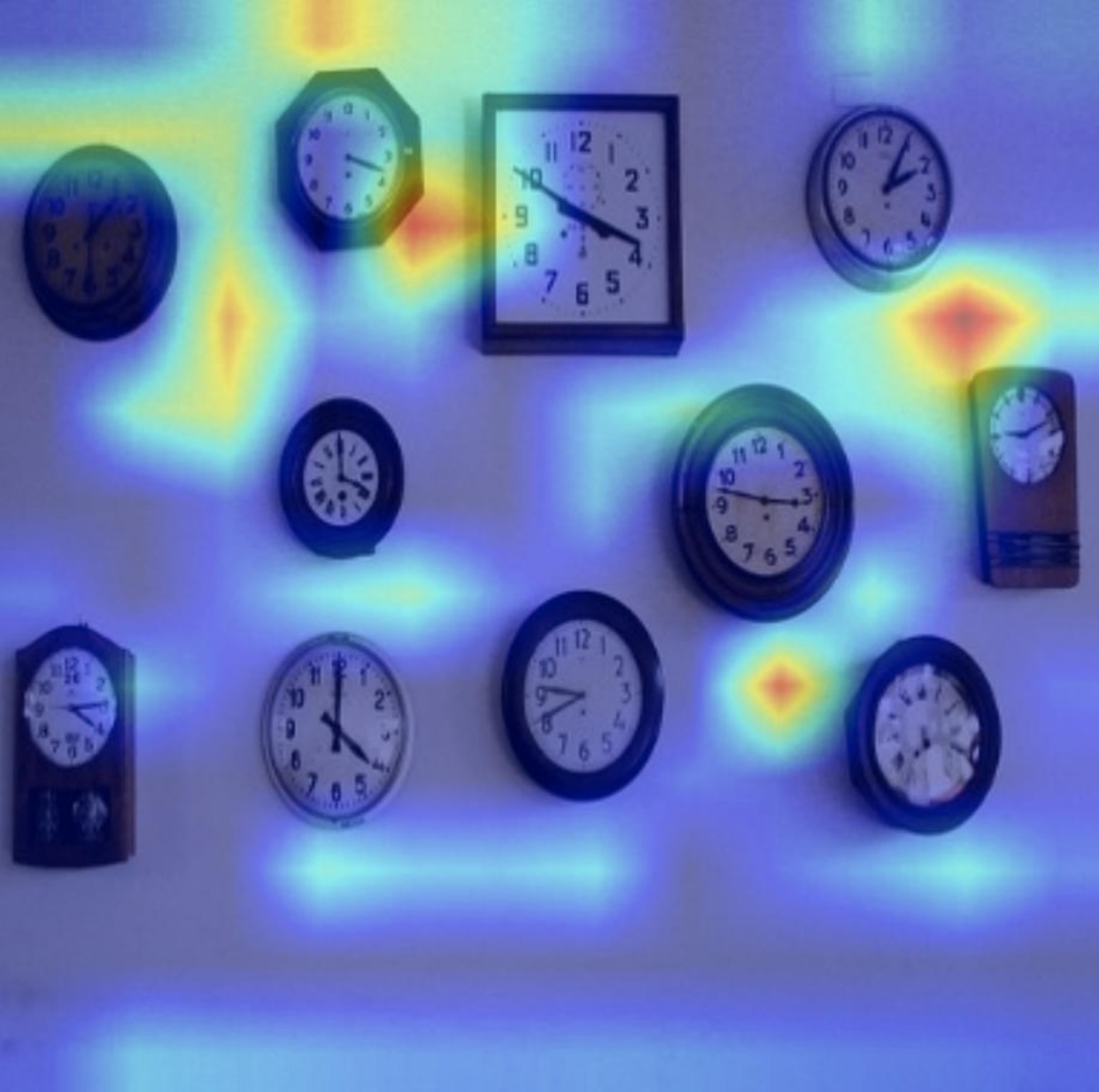}
        \end{minipage}
        &
        \begin{minipage}{0.125\hsize}
            \centering
            \includegraphics[height=16mm, width=16mm, keepaspectratio=false]{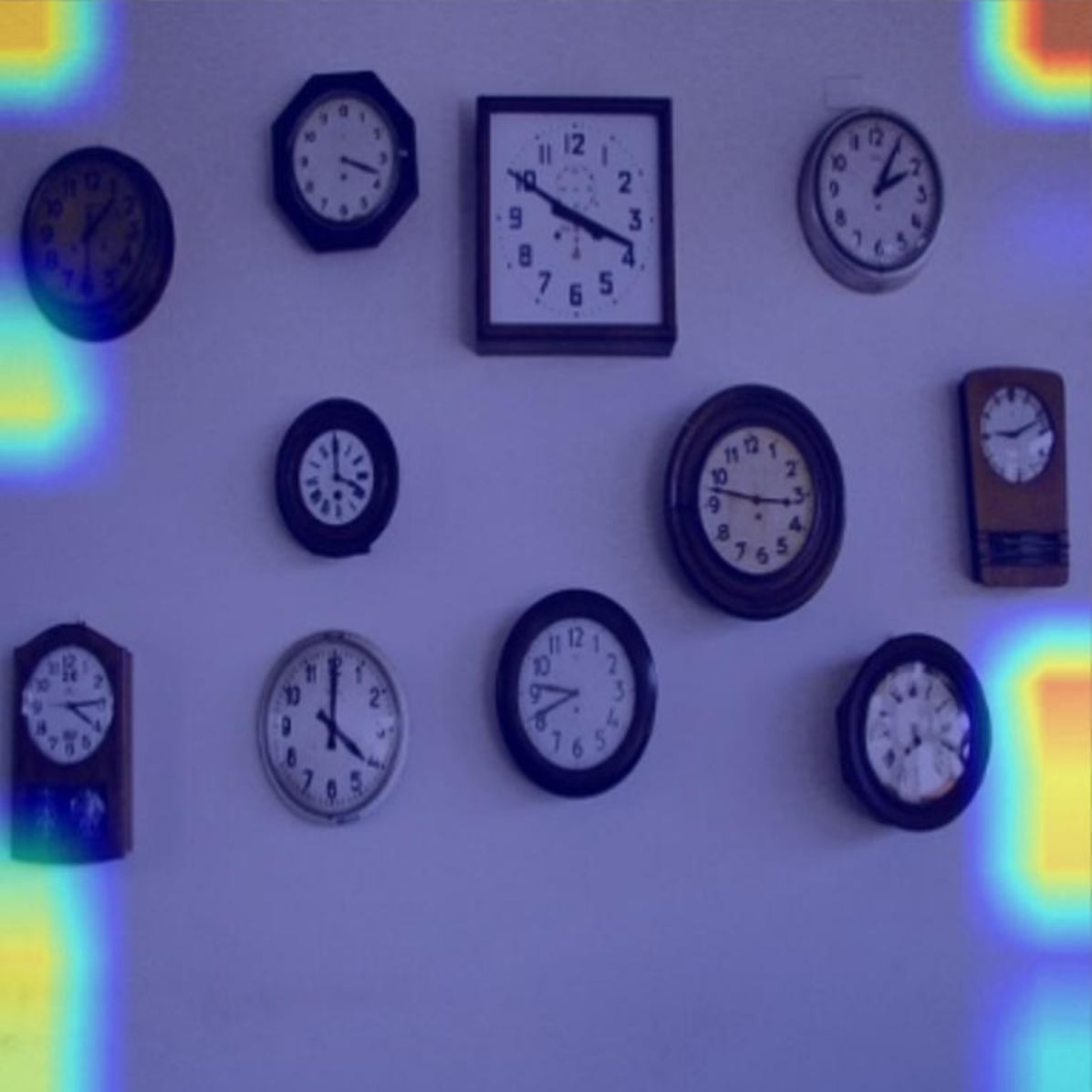}
        \end{minipage}
         \begin{minipage}{0.125\hsize}
            \centering
            \includegraphics[height=16mm, width=16mm, keepaspectratio=false]{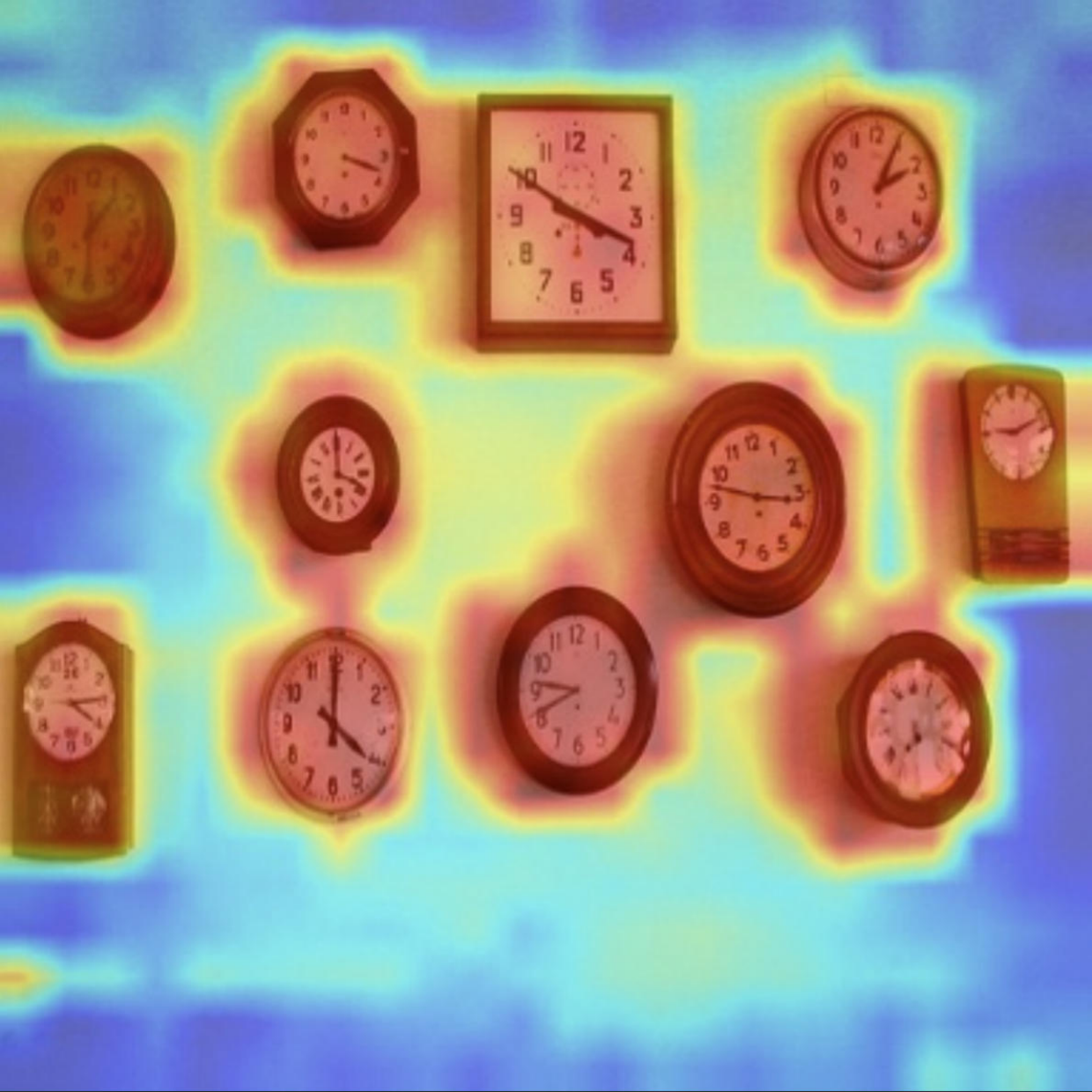}
        \end{minipage}
    \end{tabular} \\
    \begin{tabular}{cccccccc}
        \begin{minipage}{0.125\hsize}
            \centering
            \includegraphics[height=16mm, width=16mm, keepaspectratio=false]{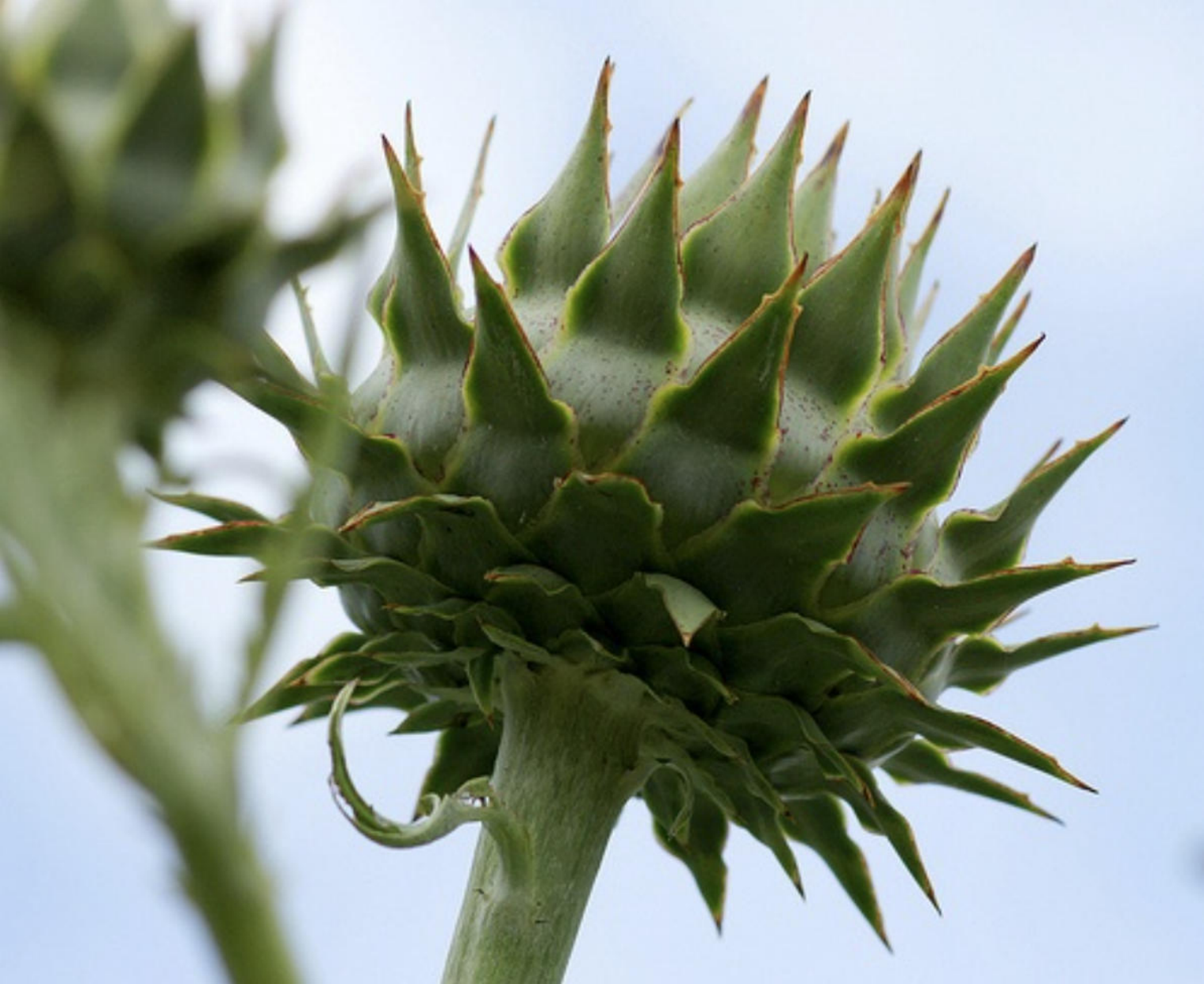}
        \end{minipage}
         &  
        \begin{minipage}{0.125\hsize}
            \centering
            \includegraphics[height=16mm, width=16mm, keepaspectratio=false]{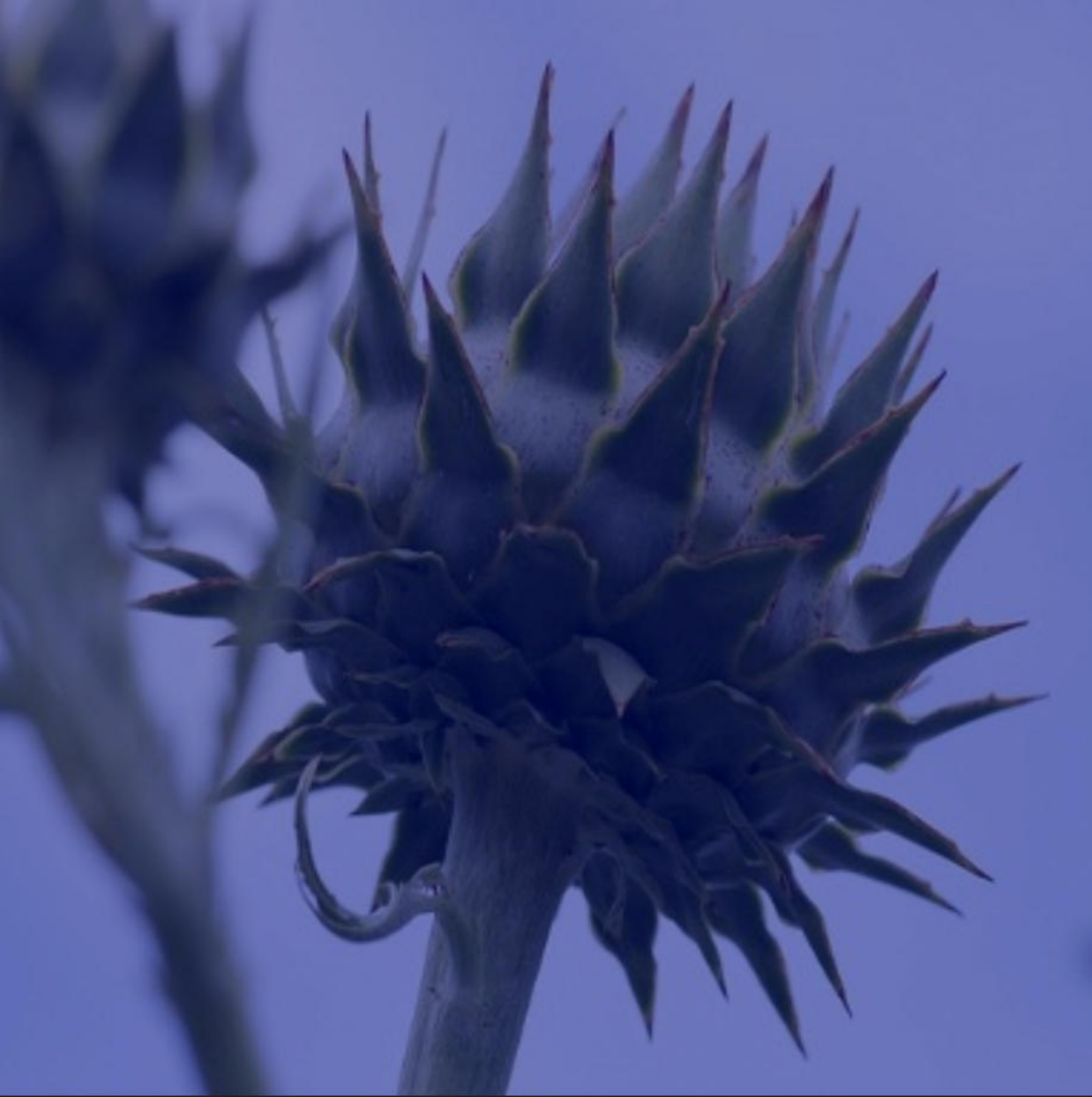}
        \end{minipage}
         & 
         \begin{minipage}{0.125\hsize}
            \centering
            \includegraphics[height=16mm, width=16mm, keepaspectratio=false]{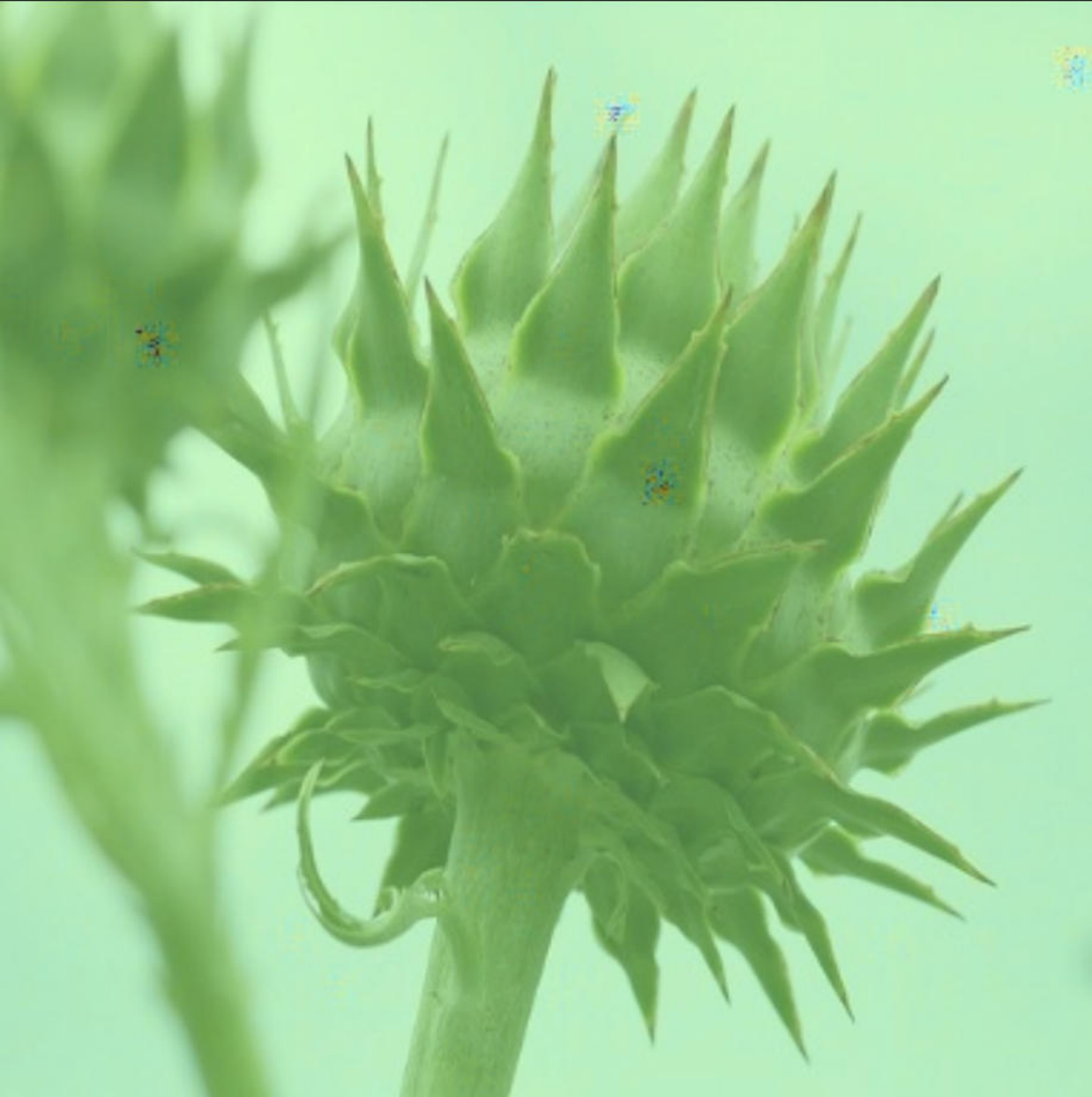}
        \end{minipage}
        &
         \begin{minipage}{0.125\hsize}
            \centering
            \includegraphics[height=16mm, width=16mm, keepaspectratio=false]{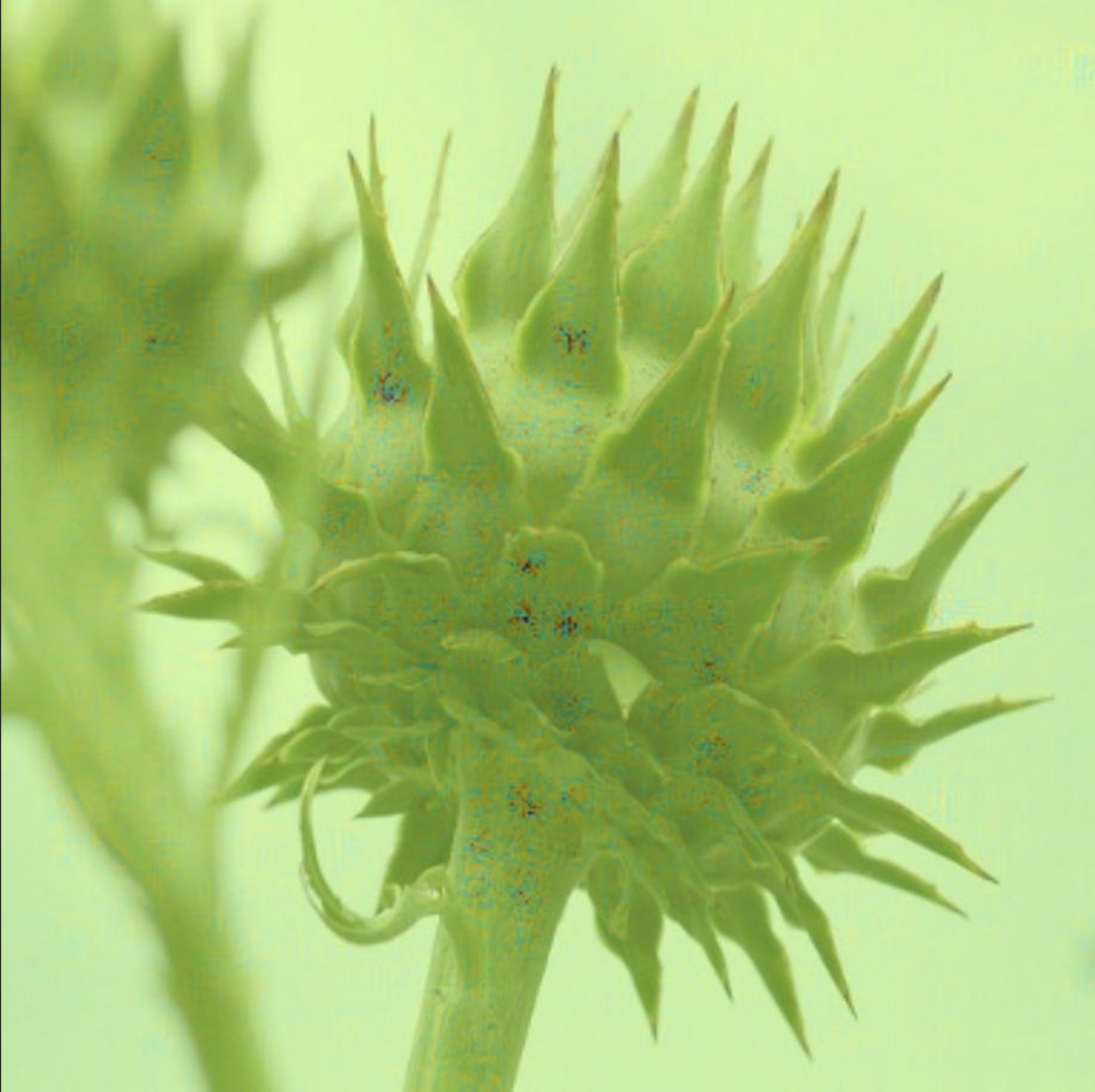}
        \end{minipage}
        &
         \begin{minipage}{0.125\hsize}
            \centering
            \includegraphics[height=16mm, width=16mm, keepaspectratio=false]{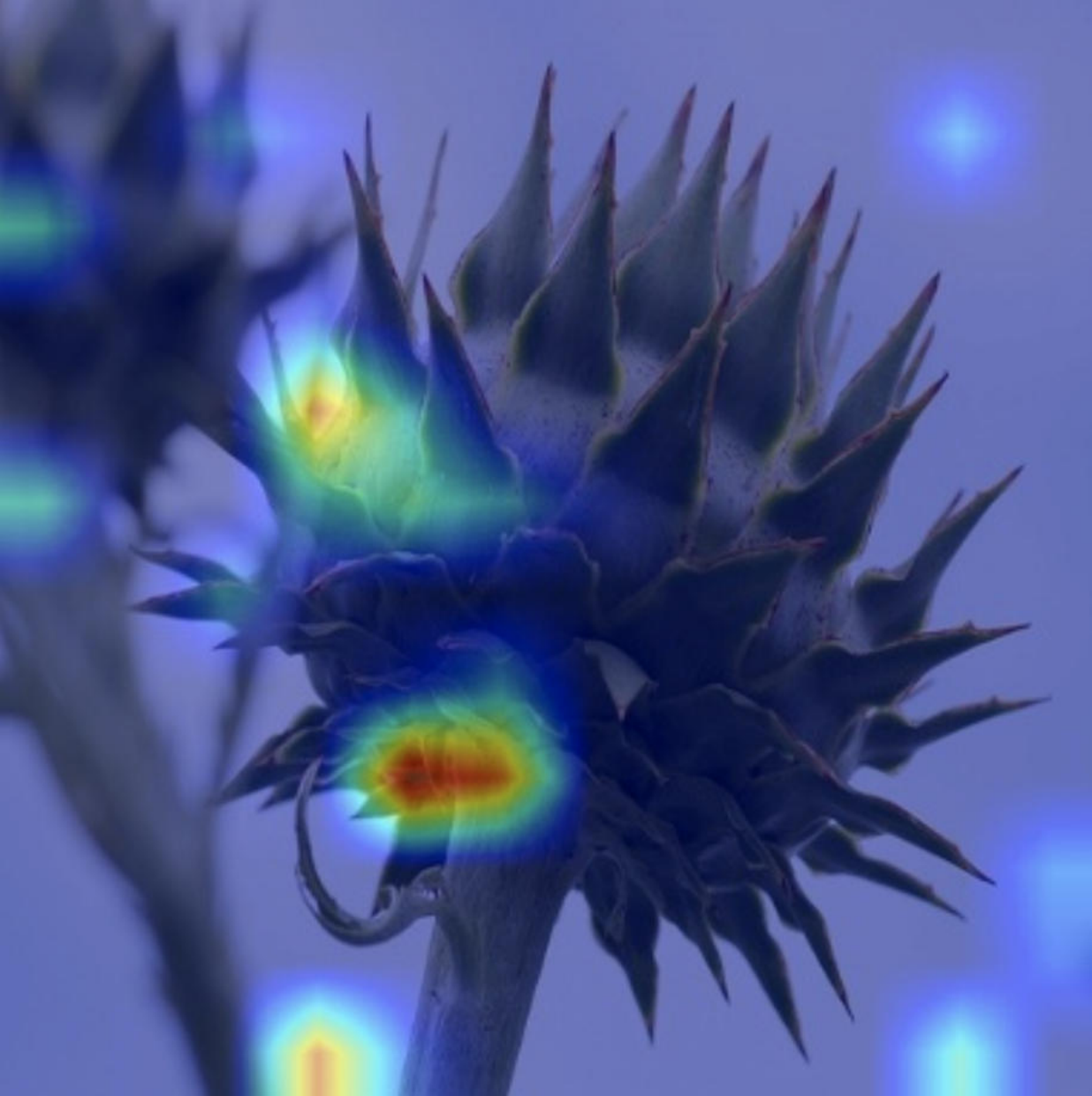}
        \end{minipage}
        &
        \begin{minipage}{0.125\hsize}
            \centering
            \includegraphics[height=16mm, width=16mm, keepaspectratio=false]{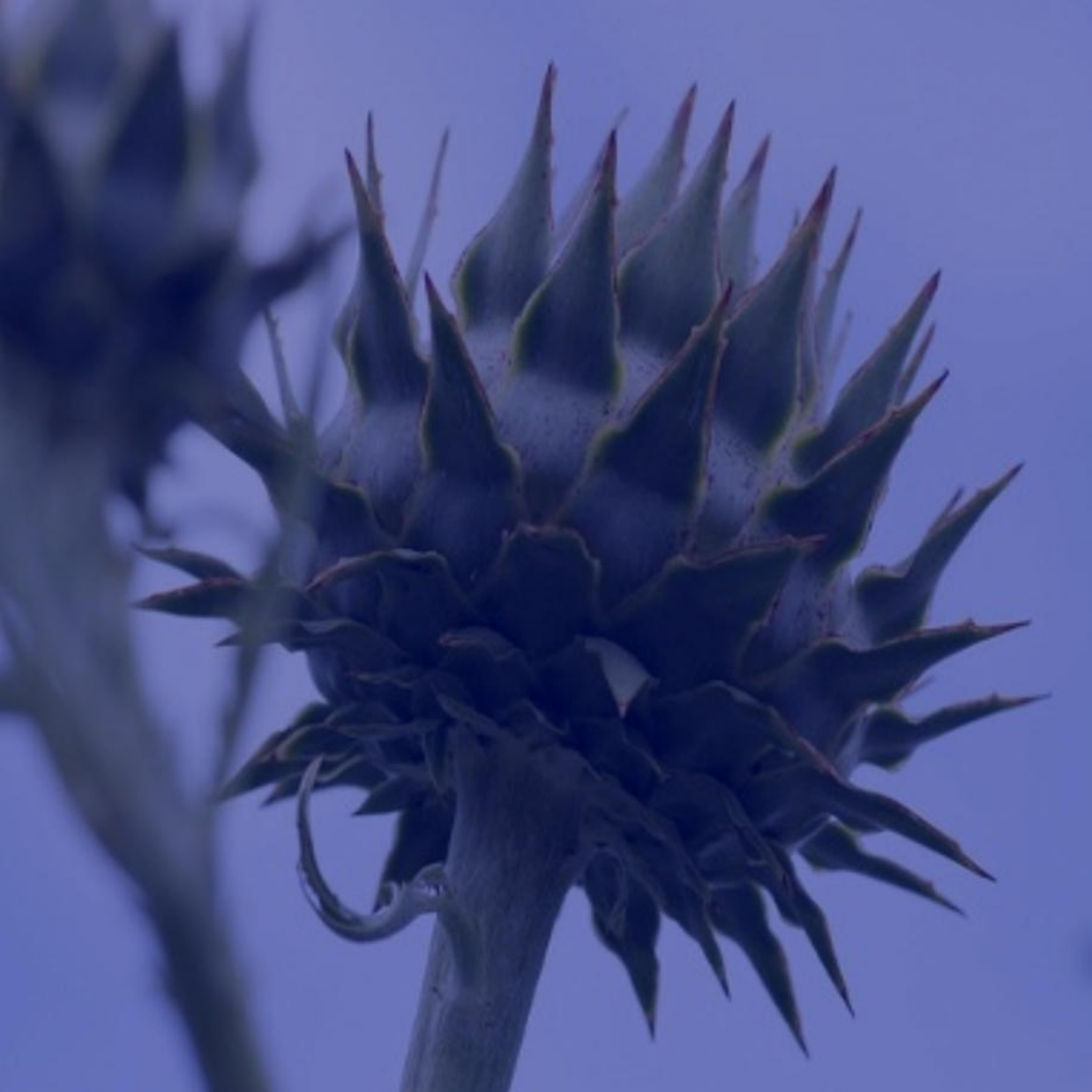}
        \end{minipage}
         \begin{minipage}{0.125\hsize}
            \centering
            \includegraphics[height=16mm, width=16mm, keepaspectratio=false]{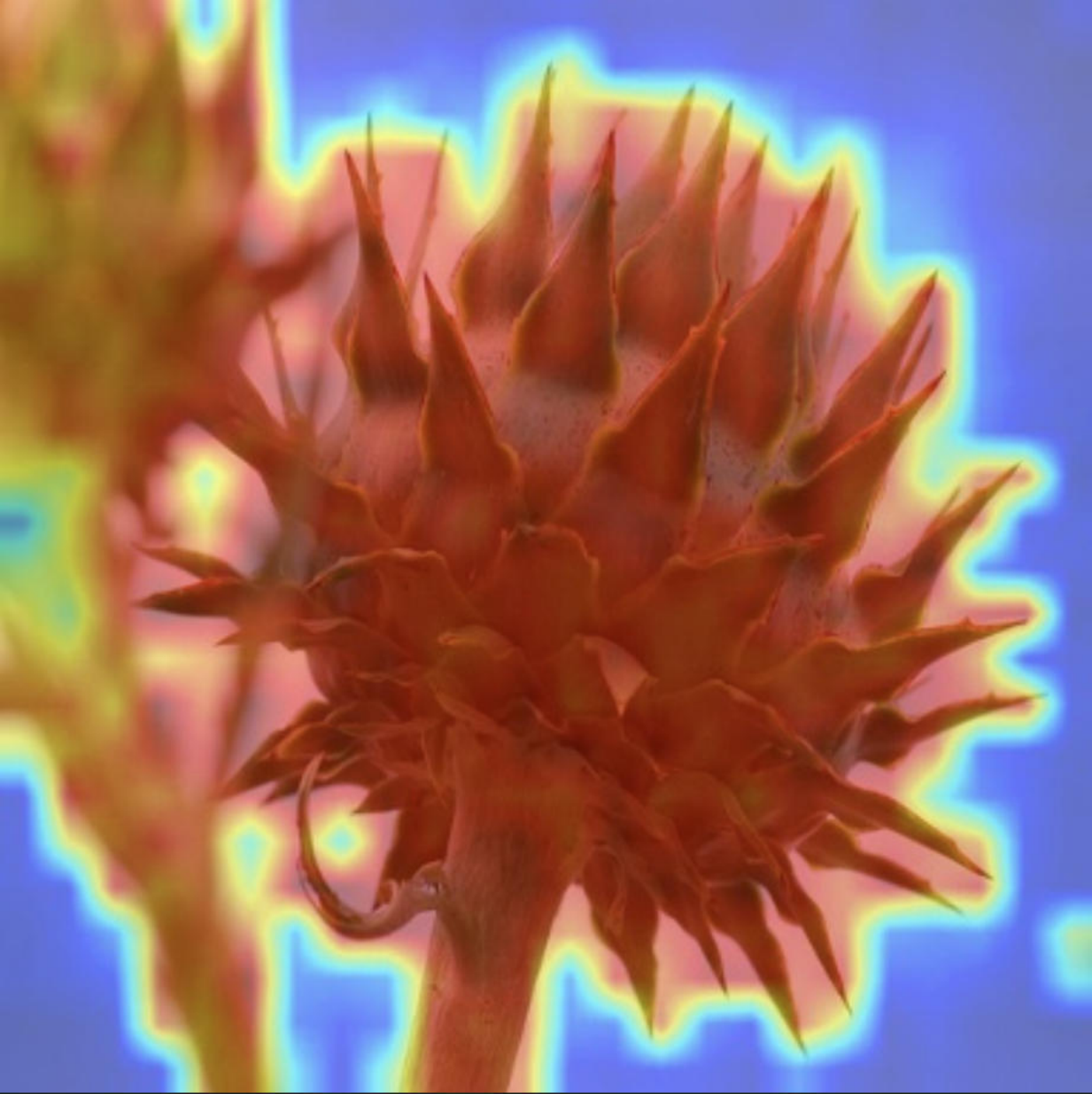}
        \end{minipage}
    \end{tabular} \\ \vspace{1mm}
    \begin{tabular}{cccccccc}
        \begin{minipage}{0.125\hsize}
            \centering
            \includegraphics[height=16mm, width=16mm, keepaspectratio=false]{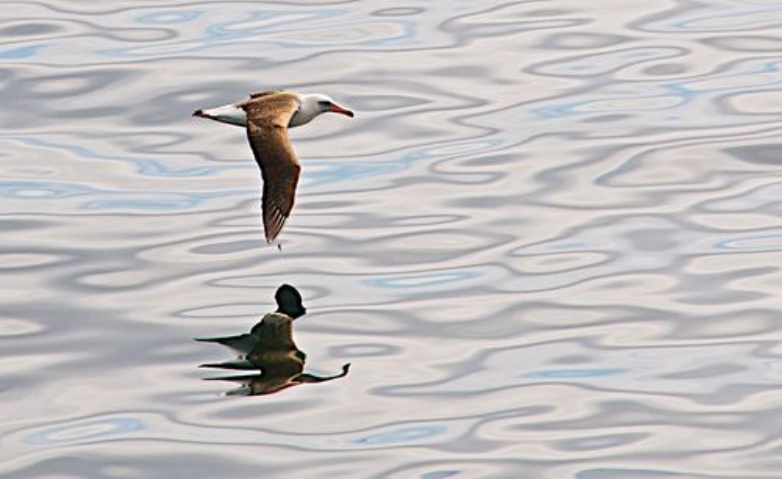}
        \end{minipage}
         &  
        \begin{minipage}{0.125\hsize}
            \centering
            \includegraphics[height=16mm, width=16mm, keepaspectratio=false]{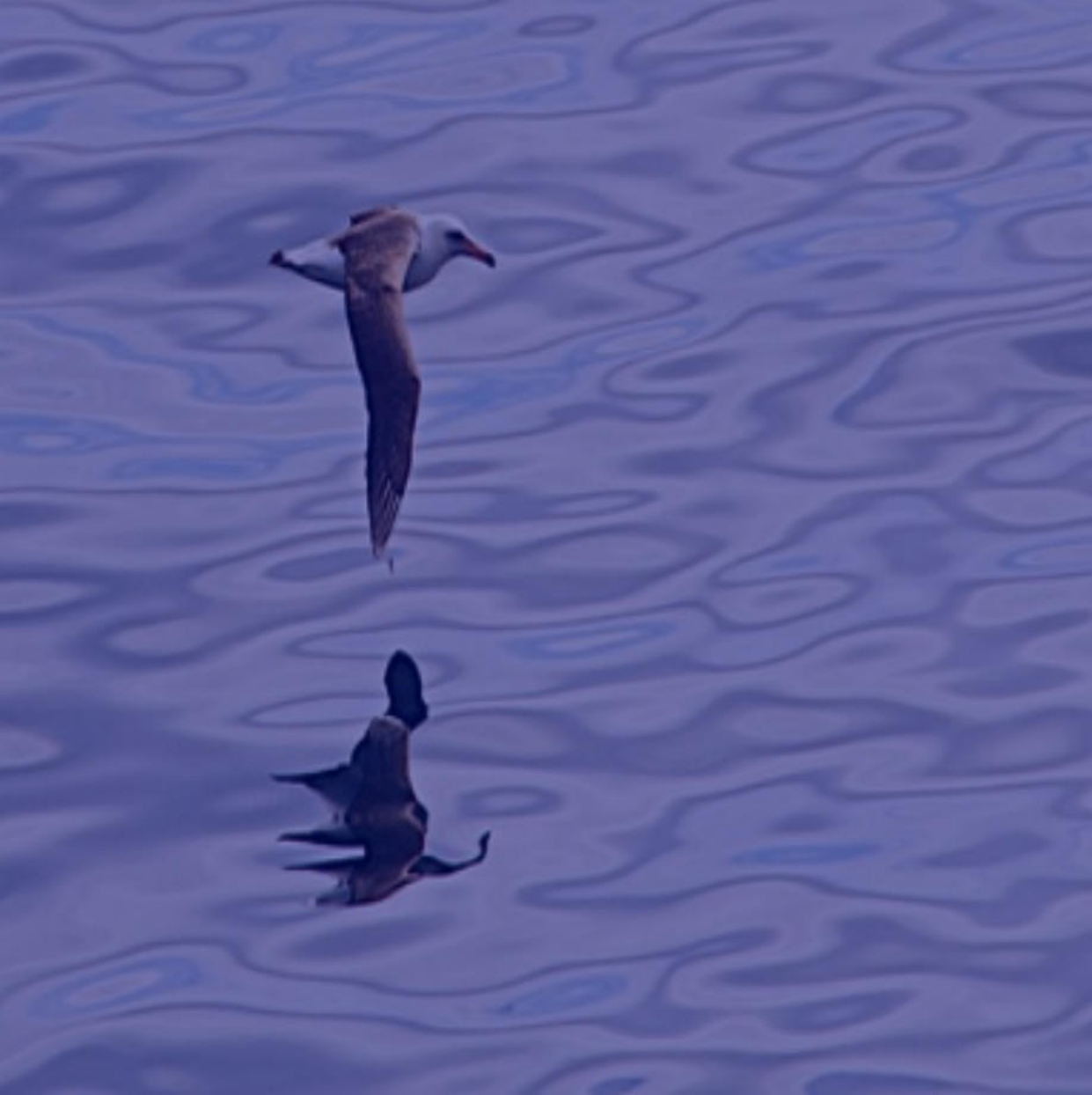}
        \end{minipage}
         & 
         \begin{minipage}{0.125\hsize}
            \centering
            \includegraphics[height=16mm, width=16mm, keepaspectratio=false]{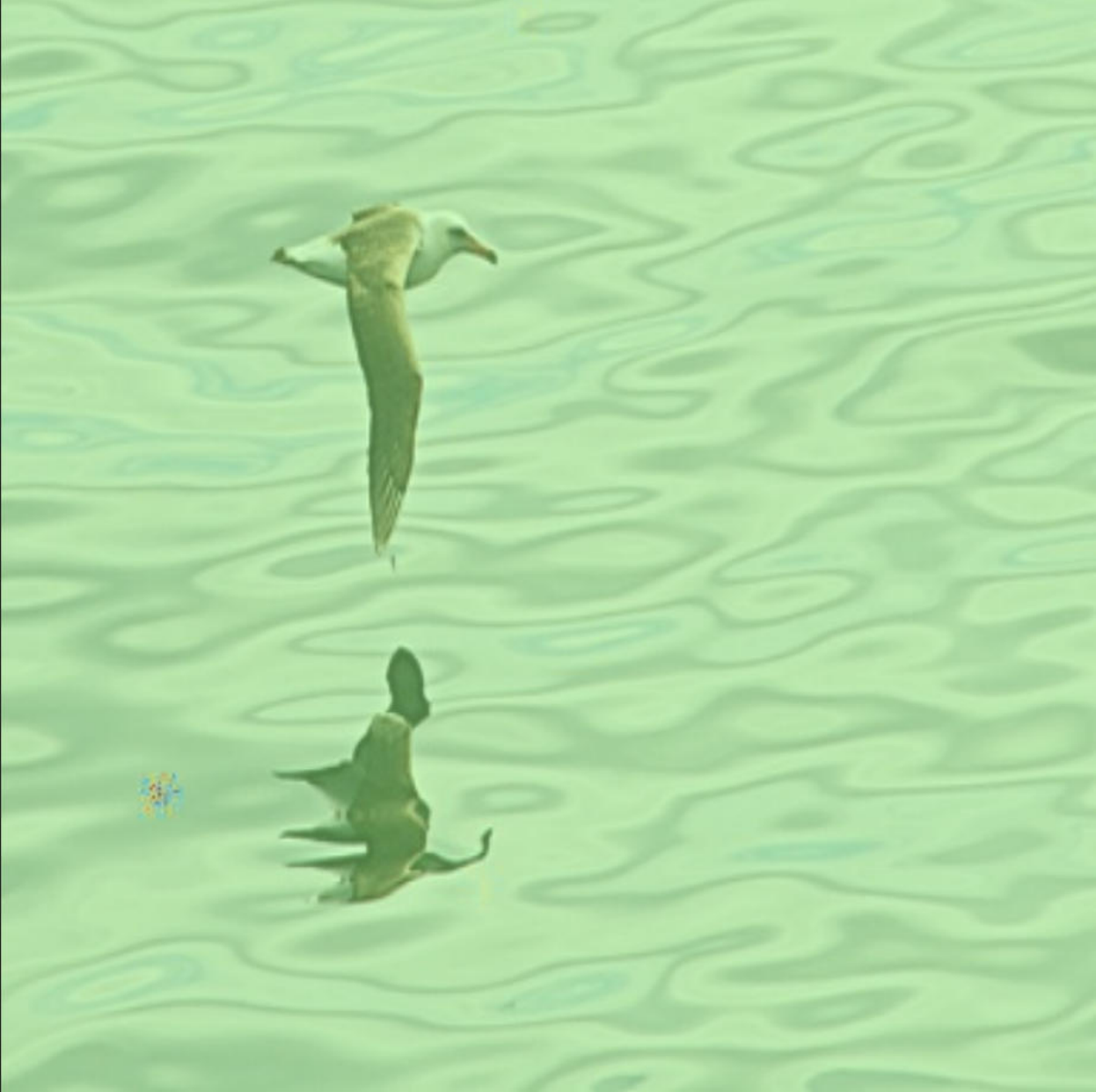}
        \end{minipage}
        &
         \begin{minipage}{0.125\hsize}
            \centering
            \includegraphics[height=16mm, width=16mm, keepaspectratio=false]{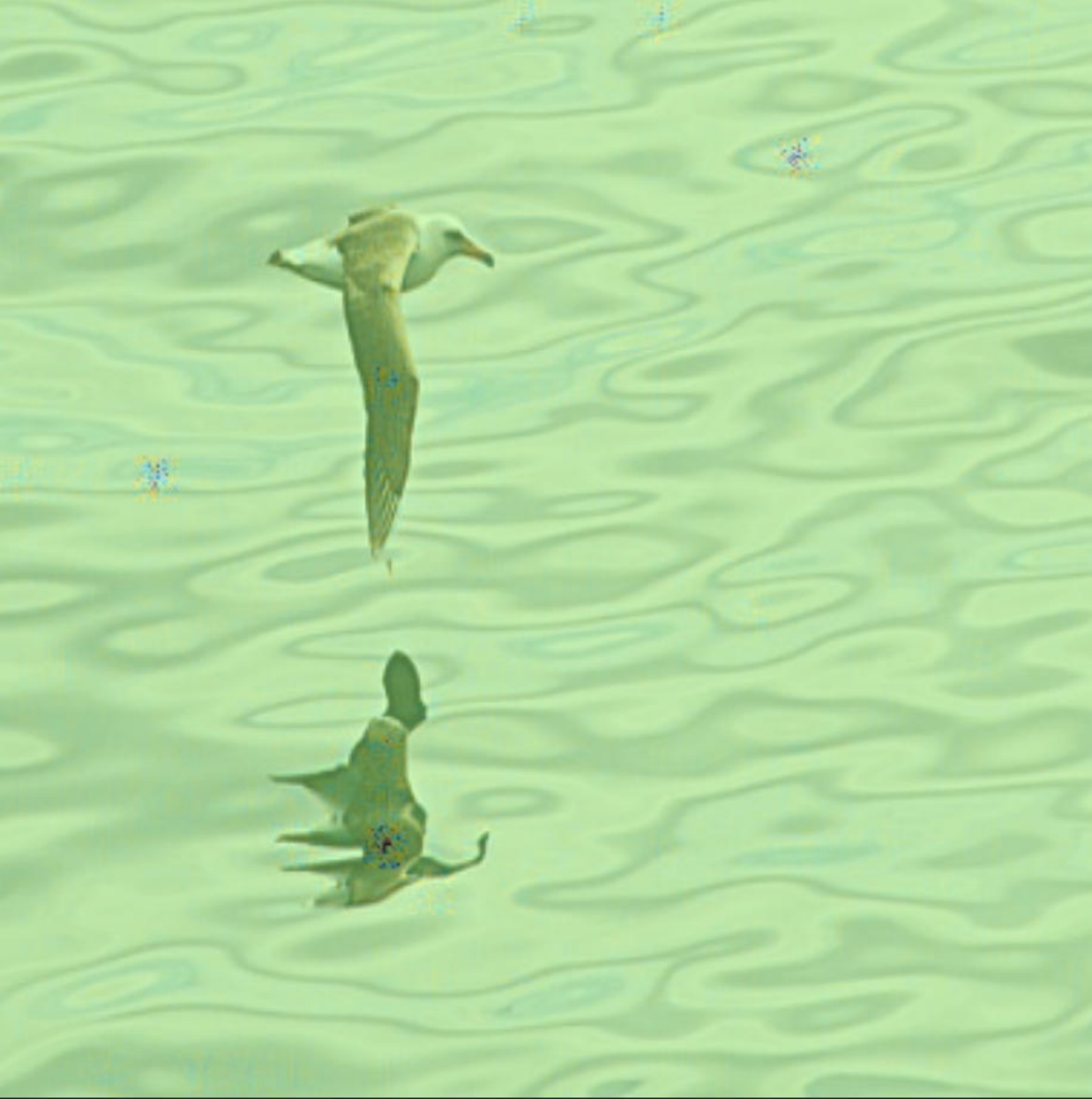}
        \end{minipage}
        &
         \begin{minipage}{0.125\hsize}
            \centering
            \includegraphics[height=16mm, width=16mm, keepaspectratio=false]{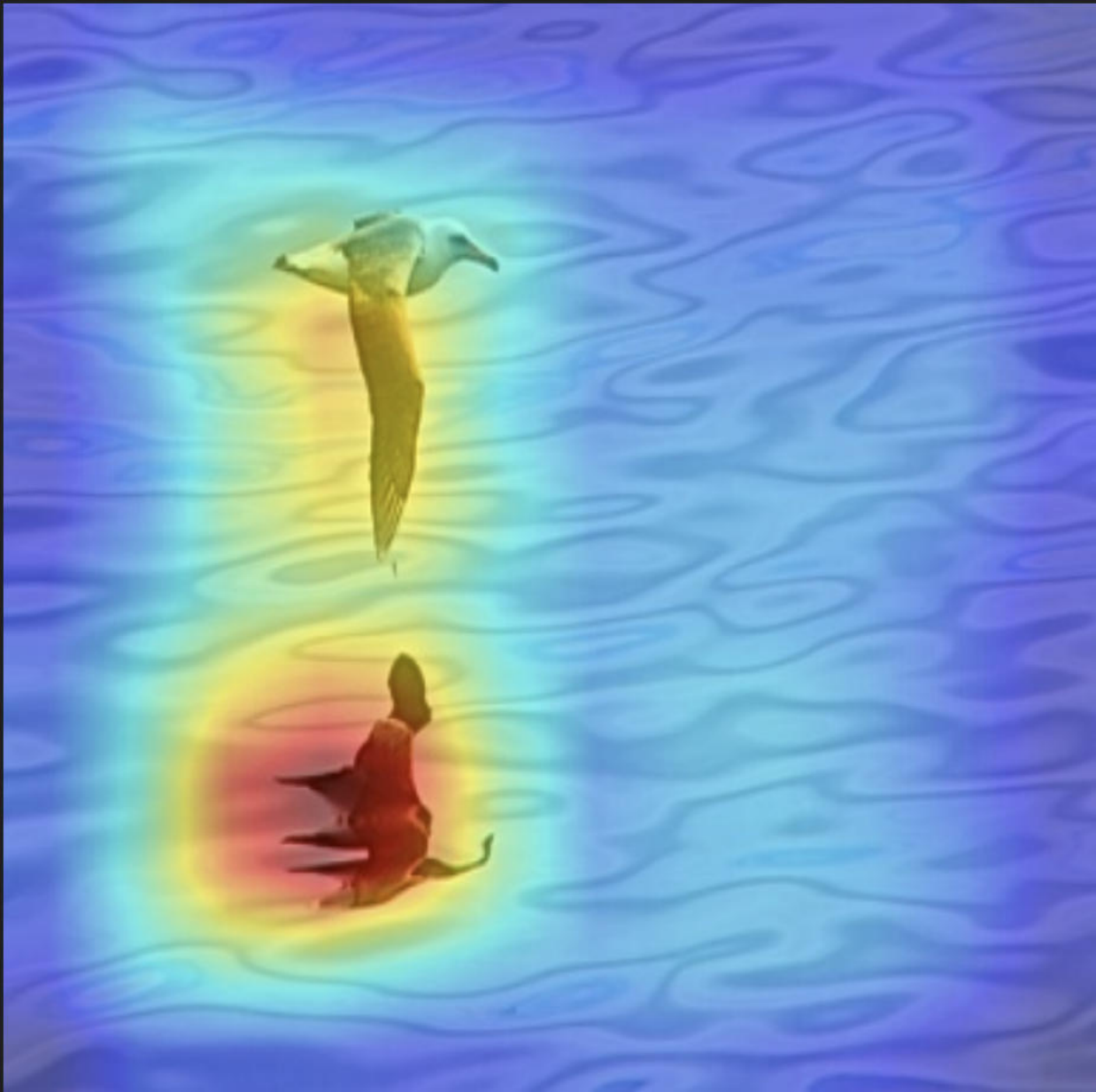}
        \end{minipage}
        &
        \begin{minipage}{0.125\hsize}
            \centering
            \includegraphics[height=16mm, width=16mm, keepaspectratio=false]{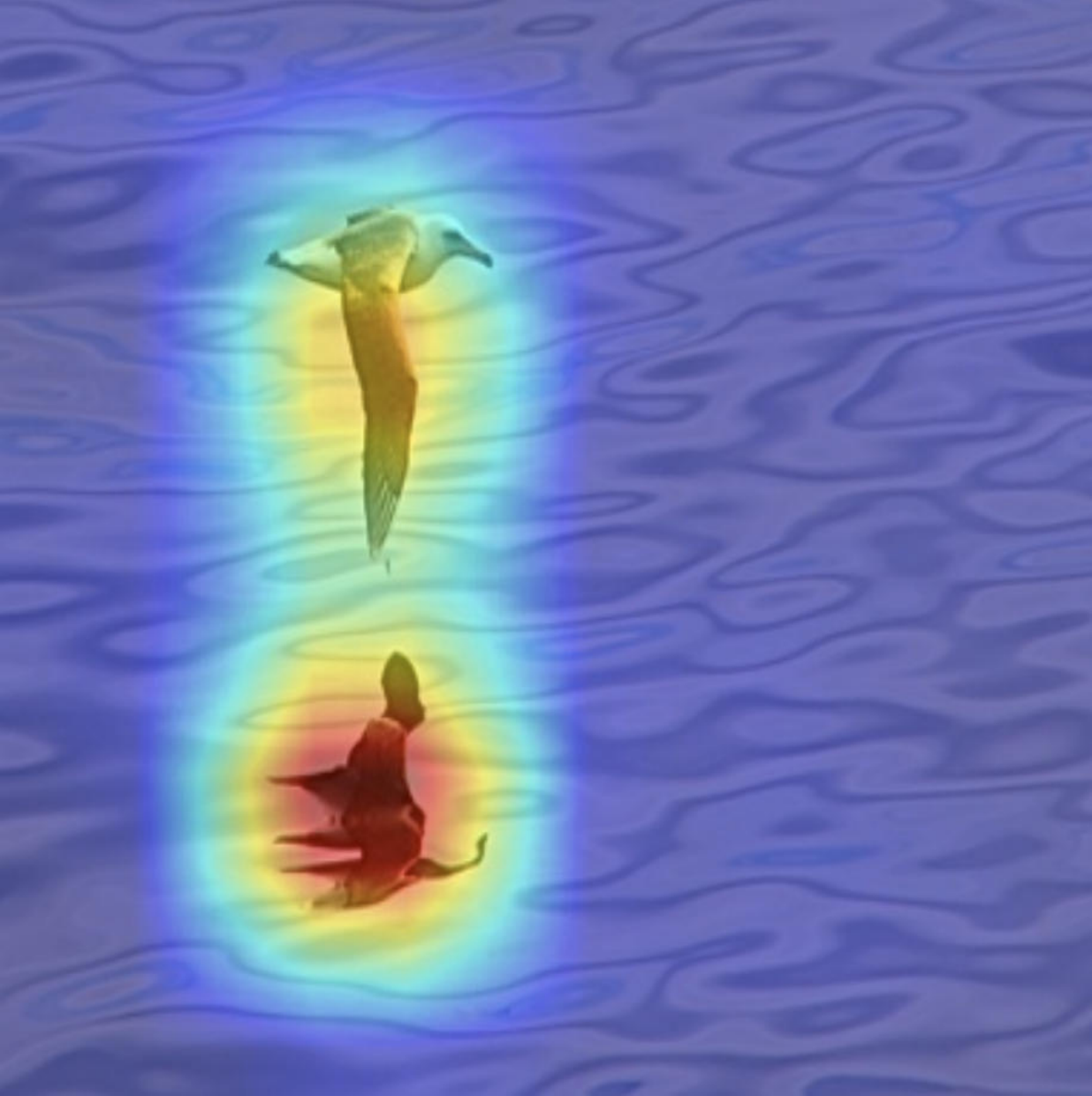}
        \end{minipage}
         \begin{minipage}{0.125\hsize}
            \centering
            \includegraphics[height=16mm, width=16mm, keepaspectratio=false]{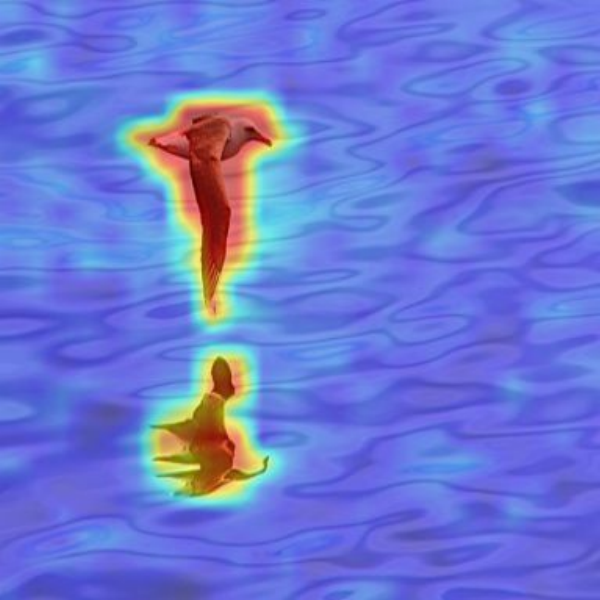}
        \end{minipage}
    \end{tabular}
    \centering
    \begin{tabular}{ccccccc}
        \begin{minipage}{0.125\hsize}
            \centering
            \scriptsize{(a) Original}
        \end{minipage}
         &  
        \begin{minipage}{0.125\hsize}
            \centering
            \scriptsize{(b) LRP}
        \end{minipage}
         & 
         \begin{minipage}{0.125\hsize}
            \centering
            \scriptsize{(c) Integrated gradients}
        \end{minipage}
        &
         \begin{minipage}{0.125\hsize}
            \centering
            \scriptsize{(d) \\ Guided\,BP}
        \end{minipage}
        &
         \begin{minipage}{0.125\hsize}
            \centering
            \scriptsize{(e) Grad-CAM}
        \end{minipage}
        &
        \begin{minipage}{0.125\hsize}
            \centering
            \scriptsize{(f) Score-CAM}
        \end{minipage}
         \begin{minipage}{0.125\hsize}
            \centering
            \scriptsize{(g) \\ Ours}
        \end{minipage}
    \end{tabular} 
    \caption{Qualitative results of successful and failure cases. The first to fourth rows display successful examples, whereas the fifth row illustrates a failure case. Column (a) depicts the original images. Columns (b)-(f) present the explanations generated by baseline methods, overlaid on the original images. By contrast, column (g) displays the visual explanation produced by our proposed method. Notably, as evident in the fifth row, our method generated attention maps that appropriately focused on relevant regions.}
    \label{fig:qualitative}
\end{figure}

\label{sec:qualitative}
Fig. \ref{fig:qualitative} shows the qualitative results of successful and failure cases. Column (a) displays the original images, whereas columns (b)-(f) show the explanations generated by the baseline methods overlaid on the original images, and column (g) represents the results generated by the proposed method.

Columns (b), (c), and (d) show the explanations generated by LRP, IG, and Guided BP, respectively. 
These results show that these methods were suboptimal, because they tended to assign uniform attention across most regions.
Columns (e) and (f) show the explanations generated by Grad-CAM and Score-CAM, respectively.
Although both methods focused on portions of the target object, they also erroneously attended to irrelevant areas.
By contrast, the attention maps generated by our proposed method distinctly targeted relevant objects with detailed precision, while minimizing attention to background regions. This indicates that our method provided more appropriate and relevant explanations.

The fifth row of Fig. \ref{fig:qualitative} shows an instance where our proposed method failed to generate appropriate explanations. Specifically, columns (e)-(g) display the attribution maps generated by Grad-CAM, Score-CAM, and our proposed method, respectively. These attention maps erroneously focused on regions where birds were reflected on the water surface, which is inappropriate for the given context. 
Furthermore, as illustrated in columns (b)-(d), LRP, IG, and Guided BP also failed to produce relevant attention maps.

\newcolumntype{Y}{>{\centering\arraybackslash}X} 

\begin{table}[t]
\centering
\caption{Quantitative results on the CUB-200-2011\cite{WahCUB_200_2011} dataset. LRP, IG and Guided BP represent Layer-wise Relevance Propagation, integrated gradients and guided backpropagation, respectively. The best records are highlighted in \textbf{bold}.}
\vspace{-2mm}
\renewcommand{\arraystretch}{1.5}
\begin{tabularx}{\linewidth}{XYYYY}
\bhline{1.25pt}
  
Method                                     & \multicolumn{1}{c}{mIoU (↑)} & \multicolumn{1}{c}{Insertion (↑)}          & \multicolumn{1}{c}{Deletion (↓)} & \multicolumn{1}{c}{ID Score (↑)} \\ \hline 
LRP \cite{bach2015layerwiserelevancepropagationlrp}                            & 0.000{\scriptsize ±0.000}                & 0.215{\scriptsize ±0.021}                              & 0.010{\scriptsize ±0.004}                    & 0.211{\scriptsize ±0.025}                    \\ 
IG \cite{sundararajan2017axiomatic}   & 0.038{\scriptsize ±0.004}                & 0.415{\scriptsize ±0.031}                              & 0.022{\scriptsize ±0.003}                    & 0.393{\scriptsize ±0.029}                    \\ 
\begin{tabular}{c}Guided  BP \cite{springenberg2015striving} \end{tabular} & 0.062{\scriptsize ±0.005}                & 0.287{\scriptsize ±0.015}                              & 0.043{\scriptsize ±0.003}                    & 0.244{\scriptsize ±0.013}                    \\ 
Score-CAM \cite{wangScoreCAMScoreWeightedVisual2020}                      & 0.160{\scriptsize ±0.032}                & 0.604{\scriptsize ±0.044}                              & 0.091{\scriptsize ±0.026}                    & 0.514{\scriptsize ±0.068}                    \\ 
Grad-CAM \cite{Selvaraju2017gradcam}                  & 0.161{\scriptsize ±0.026}                & 0.430{\scriptsize ±0.027}                              & 0.209{\scriptsize ±0.024}                    & 0.221{\scriptsize ±0.037}                    \\ 
Ours                                       & \textbf{0.693{\scriptsize ±0.007}}       &  \textbf{0.704{\scriptsize ±0.012}} & \textbf{0.007{\scriptsize ±0.002}}           & \textbf{0.697{\scriptsize ±0.011}}           \\ \bhline{1.25pt}
\end{tabularx}
\label{CUB_quantative_result}

\end{table}

\begin{table}[t]
\centering
\caption{Quantitative results on the ImageNet-S\cite{gao2022luss} dataset. LRP, IG and Guided BP represent Layer-wise Relevance Propagation, integrated gradients and guided backpropagation, respectively. The highest-performing records are marked in \textbf{bold}.}
\vspace{-2mm}
\renewcommand{\arraystretch}{1.5}

\begin{tabularx}{\linewidth}{XYYYY}

\bhline{1.25PT}
Method                                   & mIoU (↑) & Insertion (↑) & Deletion (↓) & ID Score (↑) \\ \hline 
LRP \cite{bach2015layerwiserelevancepropagationlrp}                                                  & 0.007                                              & 0.314                                                   & 0.006                          & 0.298                          \\ 
IG \cite{sundararajan2017axiomatic} & 0.215            & 0.098                                          & 0.007                          & 0.091                          \\ 
Guided BP \cite{springenberg2015striving}                       & 0.215                                              & 0.107                                                   & 0.010                          & 0.097                          \\ 
Score-CAM \cite{wangScoreCAMScoreWeightedVisual2020}                                            & 0.200                                              & 0.377                                                   & 0.033                          & 0.344                          \\ 
Grad-CAM \cite{Selvaraju2017gradcam}                                        & 0.218                                              & 0.186                                                   & 0.060                           & 0.126                          \\ 
Ours                                                             & \textbf{0.379}                                     & \textbf{0.462}                                          & \textbf{0.003}                 & \textbf{0.459}                 \\ \bhline{1.25pt}
\end{tabularx}

\label{ImageNet-S_quantative_result}
\end{table}

\vspace{-2mm}
\subsection{Quantitative Results} 
\vspace{-2mm}

Tables \ref{CUB_quantative_result} and \ref{ImageNet-S_quantative_result} present the quantitative results of the comparison of the proposed method with the baseline methods. On the CUB-200-2011 dataset, we conducted five trials for each method and reported their mean and standard deviation. By contrast, for the ImageNet-S dataset, we conducted a single trial for each method.
In this experiment, we used several evaluation metrics: mean IoU, insertion score, deletion score, and the insertion-deletion score (ID Score). Among these, we designated mean IoU as the primary metric for evaluation.
The mean IoU is defined as follows:
\begin{equation*}
    \mathrm{mean \ IoU} = \frac{1}{N}\sum^N_{i=1} \mathrm{IoU}(\hat{\bm{y}_i}, \bm{y}_i),
\end{equation*}
where $N$ denotes the number of samples. $\bm{\hat{y_i}}$ and $\bm{y_i}$ denote the predicted and ground-truth masks in the $i$-th sample, respectively. IoU denotes the IoU between two masks. 
The insertion score and deletion score are calculated as the area under curve (AUC) of the insertion and deletion curves, respectively. The ID Score is then defined as the difference between the insertion score and deletion score.
The insertion and deletion curves represent the changes in prediction when important regions based on $\bm{\alpha}$ are inserted or deleted, respectively. The computation of these scores is defined as follows:
First, we sort the elements of $\bm{\alpha}$ in descending order: $\alpha_{i_1,j_1}, \alpha_{i_2,j_2}, …, \alpha_{i_w, i_h}$. We then define the sets $A_n, \bm{i}_n,$ and $\bm{d}_n$ as follows:
\begin{equation*}
    A_n = \{(i_k, j_k) | k \leq n\},
\end{equation*}
\begin{equation*}
(i_n, d_n) = \begin{cases}
(x_{ij}, 0) & \text{if } (i, j) \in A_n, \\
(0, x_{ij}) & \text{if } (i, j) \notin A_n.
\end{cases}
\end{equation*}
where $n$ represents the number of pixels to be inserted or deleted. 
We input $\bm{i}_n$ and $\bm{d}_n$ into the model. The resulting outputs are denoted by $\bm{y}^{(\mathrm{ins}, n)}$ and $\bm{y}^{(\mathrm{del}, n)}$, respectively. The curves plotted against $n$, that is $y^{(\mathrm{ins}, n)}_C$ and $y^{(\mathrm{del}, n)}_C$, are the insertion and deletion curves for a given class $C$, to which the instance $\bm{x}$ belongs.

As shown in Tables \ref{CUB_quantative_result} and \ref{ImageNet-S_quantative_result}, in the CUB dataset experiments, our method achieved a mean IoU of 0.693, whereas the mean IoU of LRP, IG, Guided BP, Score-CAM, and Grad-CAM were 0.000, 0.038, 0.062, 0.160 and 0.161, respectively. 
The proposed method outperformed the highest of the baselines, Grad-CAM, by 0.532 in mean IoU and achieved the best performance for insertion, deletion and ID scores.
Furthermore, in the ImageNet-S experiments, our method also outperformed all baselines in therms of mean IoU. Specifically, it exceeded the highest baseline, Grad-CAM, by 0.161 in terms of mean IoU.
In the CUB-200-2011 dataset, the performance difference across all metrics was statistically significant ($p < 0.05$).

\vspace{-2mm}
\subsection{Ablation Study}
\vspace{-2mm}

\begin{figure}[t]
    \centering
    \begin{tabular}{cccccccc}
        \begin{minipage}{0.4\hsize}
            \centering
            \includegraphics[height=30mm]{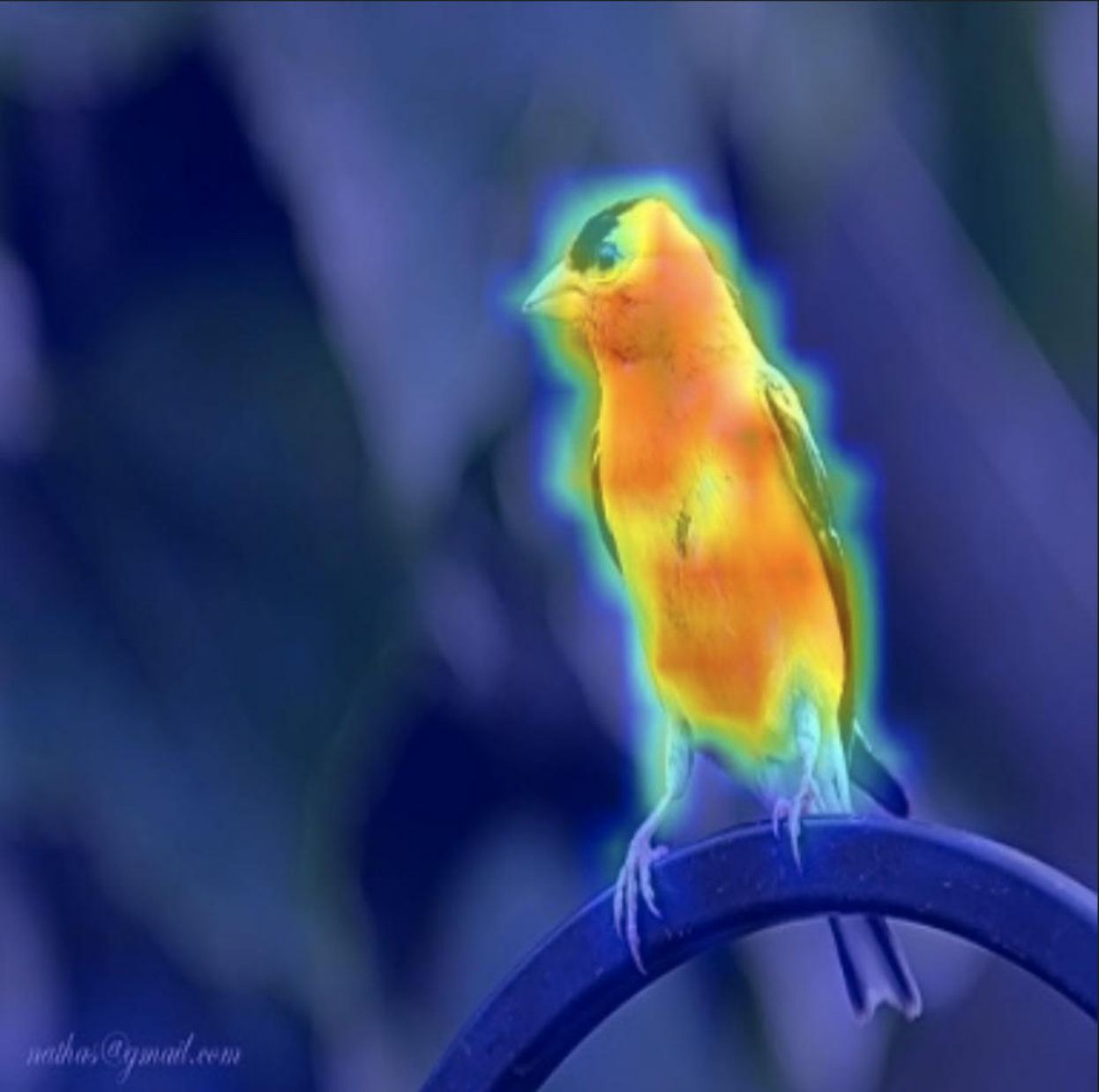}
        \end{minipage}
         &  
        \begin{minipage}{0.4\hsize}
            \centering
            \includegraphics[height=30mm]{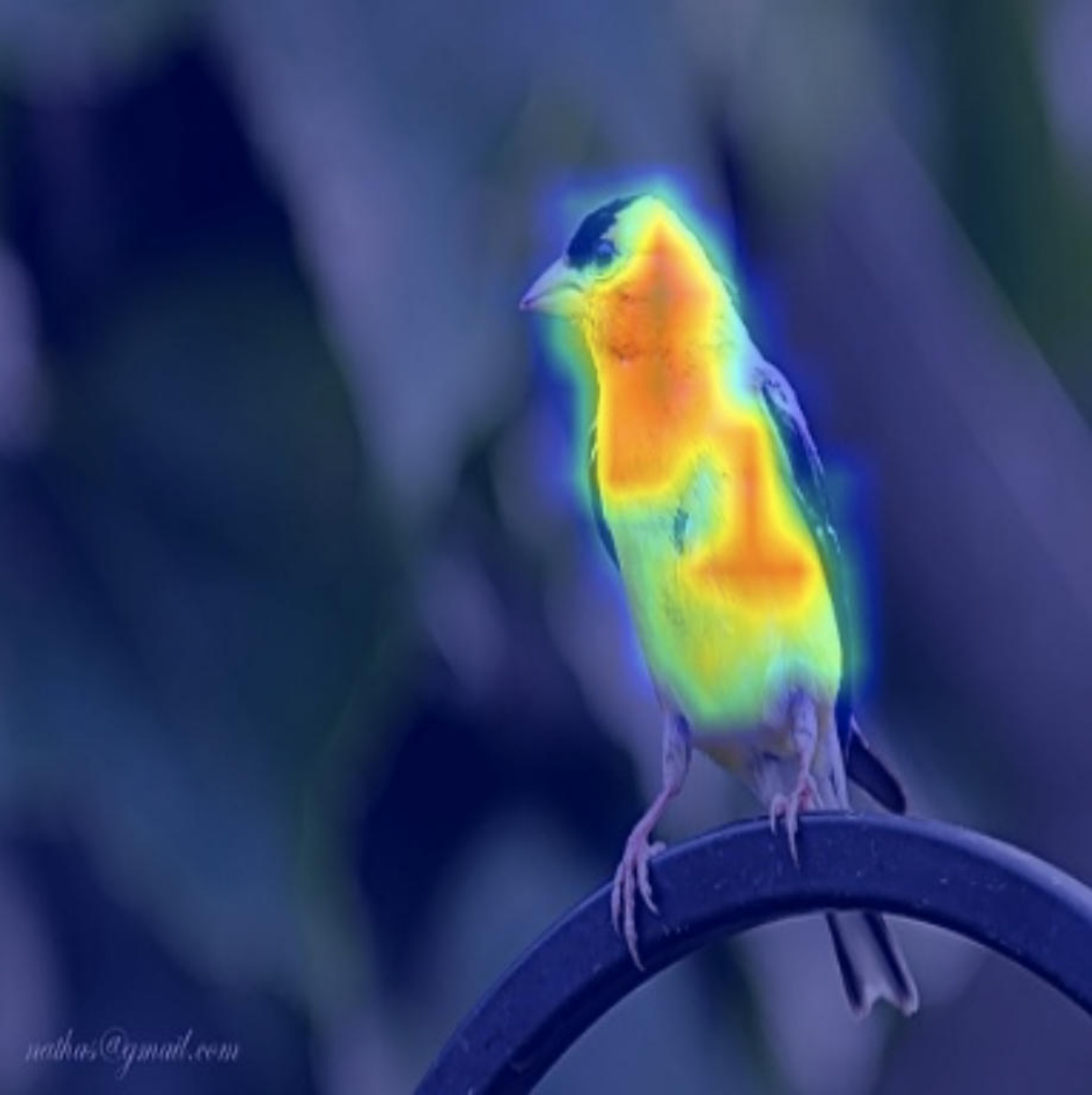}
        \end{minipage}
    \end{tabular}
    \caption{Qualitative examples illustrating the impact of AEA: with AEA (left) and without AEA (right). Our approach, incorporating AEA, broadened the focus to encompass the entire bird. By contrast, the absence of AEA resulted in attention predominantly localized to specific regions of the bird.}
    \label{AEA_samples}
\end{figure}
\newcolumntype{Y}{>{\centering\arraybackslash}X}

\begin{table}[t]
\caption{Comparison of model (i) with model (iv). The bold numbers represent the highest performance for each metric. The columns labelled AEA, LoRA, and ALA specify the inclusion of each respective attribute in the models.}
\vspace{-2mm}
\renewcommand{\arraystretch}{1.5}
\centering
\begin{tabularx}{\textwidth}{cccccYYY}
\bhline{1.25pt}
Model & AEA & LoRA & ALA & \hspace{5mm}mean IoU (↑)           & Insertion (↑)         & Deletion (↓)           & ID Score  (↑)          \\ \hline 
(i)   & \checkmark   & \checkmark   &      & \hspace{5mm}0.520{\scriptsize ±0.041}          & 0.678{\scriptsize ±0.020}          & 0.019{\scriptsize ±0.002}          & 0.659{\scriptsize ±0.021}          \\ 
(ii)  & \checkmark   &     & \checkmark    & \hspace{5mm}0.477{\scriptsize ±0.019}          & 0.595{\scriptsize ±0.054}          & 0.020{\scriptsize ±0.012}          & 0.575{\scriptsize ±0.045}          \\ 
(iii) &     & \checkmark   & \checkmark    & \hspace{5mm}0.495{\scriptsize ±0.008}          & \textbf{0.717{\scriptsize ±0.009}} & 0.014{\scriptsize ±0.004}          & \textbf{0.702{\scriptsize ±0.011}} \\ 
(iv)  & \checkmark  & \checkmark   & \checkmark    & \hspace{5mm}\textbf{0.693{\scriptsize ±0.007}} & 0.704{\scriptsize ±0.012}          & \textbf{0.007{\scriptsize ±0.002}} & 0.697{\scriptsize ±0.011}          \\ \bhline{1.25pt}
\end{tabularx}
\label{ablation}
\end{table}
We conducted three ablation studies to demonstrate the effectiveness of each module. Table \ref{ablation} presents the results of the ablation studies.
\vspace{-2mm}
\paragraph{\textbf{ALA ablation.}}
    We investigated the contribution of adding intermediate features by excluding their addition in the image encoder.
    Table \ref{ablation} shows that the mean IoU and ID score for Model (i) were 0.520 and 0.659, respectively, which indicates a decrease of 0.173 points and 0.042 points compared with Model (iv).
    This suggests that ALA plays a crucial role in extracting relevant features for explanation generation from the intermediate layers of the image encoder.
\vspace{-2mm}
\paragraph{\textbf{LoRA ablation.}}
We explored the significance of LoRA by omitting it from the model.
    From Table \ref{ablation}, the mean IoU and ID score for Model (ii) were 0.477 and 0.575, respectively. This result demonstrates reductions of 0.216 points in IoU and 0.126 points in the ID Score relative to Model (iv), which suggests that LoRA contributes to the generation of appropriate explanations.
\vspace{-2mm}
\paragraph{\textbf{AEA ablation.}}
    We investigated the effectiveness of AEA by excluding this mechanism.
    Fig. \ref{AEA_samples} shows examples of qualitative results with and without using AEA. The figure illustrates that, in the absence of AEA, the focus remained on localized regions of the bird. By contrast, our method with AEA generated attention to encompass the entire bird. 
    Moreover, Table \ref{ablation} demonstrated that the mean IoU for Model (iii) was 0.495, which is a decrease of 0.198 points compared with Model (iv). Additionally, the ID Score highlighted the competitive performance of Models (iii) and (iv).

\vspace{-2mm}
\section{Conclusion}
\vspace{-2mm}

\label{sec:conclusion}

In this study, we focused on the task of visualizing significant regions within an image, which serves as a visual explanation for a model's predictions.
The contributions of our study are as follows: 
\begin{itemize}
    \item[$\bullet$] We proposed a method for generating explanations by introducing Attention Lattice Adapter (ALA) to a VFM.
    \item[$\bullet$] We introduced Alternation Epoch Architect (AEA), which is a mechanism designed to constrain the size of attention regions by stopping updating the parameters of ALA every other epoch.
    \item[$\bullet$] Our method outperformed the baseline methods in terms of mean IoU, insertion score, deletion score, and ID Score on both the CUB-200-2011 and ImageNet-S datasets.
\end{itemize}

Although our method generated compelling results, our study had some limitations. 
First, we focused solely on CLIP as a VFM. Despite this, we believe that our approach is adaptable to other VFM architectures. Such an adaptation could illuminate differences in the output representations of various networks. 

A potential solution is the integration of modules that leverage edge detection techniques for precise object boundary extraction, thus filtering out less significant objects.

\noindent {\bf Acknowledgment}
\small
This work was partially supported by JSPS KAKENHI Grant Number 23H03478, JST CREST and NEDO.
\vspace{-5mm}
\bibliographystyle{splncs04}
\bibliography{main}
\vspace{-5mm}
\end{document}